\documentclass[journal]{IEEEtran}
\usepackage{cite}

\ifCLASSINFOpdf
  \usepackage[pdftex]{graphicx}
  \graphicspath{{figs/}}
\else
\fi
\usepackage{amsmath}
\usepackage{algorithmic}

\usepackage{array}

\ifCLASSOPTIONcompsoc
  \usepackage[caption=false,font=normalsize,labelfont=sf,textfont=sf]{subfig}
\else
  \usepackage[caption=false,font=footnotesize]{subfig}
\fi
\usepackage{fixltx2e}

\usepackage{stfloats}
\usepackage{url}

\usepackage{multirow}
\usepackage{todonotes}
\usepackage{hyperref}

\begin{document}
%
\title{Physics-Guided Deep Neural Networks for Power Flow Analysis}
%
%
%

\author{Xinyue~Hu,~\IEEEmembership{Student~Member,~IEEE,}
        Haoji~Hu,
        Saurabh~Verma,
        and~Zhi-Li~Zhang,~\IEEEmembership{Fellow,~IEEE}
\thanks{The full version of this paper is available at IEEE Transaction on Power Systems. \url{ https://ieeexplore.ieee.org/document/9216092}}
\thanks{This research was supported in part by NSF under grants CNS-1814322, CNS-1831140, CNS-1901103 and CNS-1952085, US DoD DTRA DTRA grant HDTRA1-14-1-0040, and an Amazon AWS ML Research Award. (Corresponding author: Zhi-Li Zhang)}
\thanks{X. Hu, H. Hu, S. Verma, and Z. Zhang are with the Department of Computer Science and Engineering, University of Minnesota, Minneapolis, MN, United States, 55455. (E-mail: hu000007@umn.edu, huxxx899@umn.edu, verma076@umn.edu, zhzhang@cs.umn.edu).}}
\maketitle

\begin{abstract}
Solving power flow (PF) equations is the basis of power flow analysis, which is important in determining the best operation of existing systems, performing security analysis, etc. 
However, PF equations can be out-of-date or even unavailable due to system dynamics and uncertainties, making traditional numerical approaches infeasible. 
To address these concerns, researchers have proposed data-driven approaches to solve the PF problem by learning the mapping rules from historical system operation data.
Nevertheless, prior data-driven approaches suffer from poor performance and generalizability, due to overly simplified assumptions of the PF problem or ignorance of physical laws governing power systems.
In this paper, we propose a physics-guided neural network to solve the PF problem, with an auxiliary task to rebuild the PF model.
By encoding different granularity of Kirchhoff's laws and system topology into the rebuilt PF model, our neural-network based PF solver is regularized by the auxiliary task and constrained by the physical laws.
The simulation results show that our physics-guided neural network methods achieve better performance and generalizability compared to existing unconstrained data-driven approaches.
Furthermore, we demonstrate that the weight matrices of the proposed neural networks embody power system physics by showing their similarities with the bus admittance matrices.

\end{abstract}

\begin{IEEEkeywords}
power flow analysis, power flow solver, data-driven analysis, physics-guided learning.
\end{IEEEkeywords}

\ifCLASSOPTIONpeerreview
 \begin{center} \bfseries EDICS Category: 3-BBND \end{center}
\fi
%
\IEEEpeerreviewmaketitle

%
%
%
%
\section{Introduction}
\IEEEPARstart{P}{ower} flow (PF) analysis aims at obtaining complete voltage angle and magnitude information for each bus in a power system, given specified loads, generator real power and voltage conditions\cite{saadat1999power}.
The obtained information can be used by system operators for a variety of tasks such as energy loss analysis, voltage control, reactive power planning, security analysis, and best operation determination.
Conventionally, PF analysis is achieved by first describing a power system as its state-space representation using a set of differential algebraic equations (DAEs)\cite{arrillaga1998ac} derived from prior knowledge such as the system topology and network parameters.
These DAEs, which are nonlinear and have no-closed form solutions, 
are then solved using iterative numerical methods such as Newton-Raphson method\cite{momoh1999review} to obtain voltage angles and magnitudes.

However, as modern power systems get more dynamic and uncertain, 
which are too complex to describe timely and precisely with DAEs,
conventional PF analysis approaches become ineffective\cite{chen2015measurement}.
For example, line profiles, corresponding to the parameters of DAEs, can be inaccurate due to aging and weather conditions, or even unavailable in some secondary distribution grids\cite{yu2017mapping}.
Distributed energy resources (DERs), corresponding to control rules in power system modeling, can be difficult to describe with DAEs due to missing information from vendors and individual owners\cite{sexauer2013phasor,guttromson2002modeling}.
Such deficiencies in power system modeling and analysis have posed not only safety threats to power systems such as blackouts\cite{liscouski2004final,ferc2012arizona},
but also challenges for adoption of DERs\cite{yu2017data}.

To address these concerns, researchers leverage the availability of massive measurements from phasor measurement units and smart meters, and have proposed various data-driven approaches to solve the PF problem.
We classify them into two categories: 
\textit{model rebuilding}\cite{yuan2016inverse, yu2017patopa, chen2016measurement, chen2013measurement} and 
\textit{solver learning}\cite{yu2017mapping,liu2018data,karami2008radial,muller2010artificial,baghaee2017three}.
\textit{Model rebuilding} focus on calibrating or rebuilding PF models by rediscovering model parameters (e.g., line parameters, bus admittance matrix, PF Jacobian matrix) from historical operation data. The rebuilt models are then solved using conventional numerical approaches.
Although \textit{model rebuilding} is intuitive and can potentially reflect changes of PF models, 
it suffers from drawbacks including high computational overhead, error accumulation\cite{liu2018data} and non-convergence\cite{nikkhajoei2006steady,baghaee2017three}.
In contrast, \textit{solver learning} focuses on modeling mapping rules between inputs and outputs of a power system based on historical measurement data\cite{yu2017mapping,liu2018data,karami2008radial,muller2010artificial,baghaee2017three}, thus sidestepping drawbacks in \textit{model rebuilding}.
Inspired by the structure of DC and AC power flow equations, 
several \textit{solver learning} approaches manually choose machine learning algorithms, such as linear regression (LR)\cite{liu2018data} and support vector regression (SVR) with polynomial kernels\cite{yu2017mapping}, to uncover the forward and inverse mapping rules between power injections and bus voltages.
However, LR overly simplifies PF models and SVR suffers from high computational overhead and scalability.
Other \textit{solver learning} approaches\cite{karami2008radial,muller2010artificial,baghaee2017three} view the outputs of the PF problem as implicit functions of the operating conditions, and employ neural networks such as multi-layer perceptron (MLP) and radial basis function neural network (RBFNN), to map the power-flow inputs to the power-flow outputs.
Since pure RBFNN\cite{karami2008radial} is subject to overfitting, 
subsequent efforts\cite{muller2010artificial,baghaee2017three} try to incorporate physical knowledge to avoid overfitting.
However, they make impractical assumptions such as the availability of accurate system parameters\cite{muller2010artificial} and PF models\cite{baghaee2017three}.

In this paper, we propose physics-guided deep neural networks to solve the PF problem under the assumption that system parameters and PF models can be inaccurate or unavailable.
The key idea is to utilize the physical knowledge of power systems to regularize neural networks (NNs) and alleviate the over-parameterization issue.
Unlike existing physics-related neural network approaches\cite{muller2010artificial,baghaee2017three} which make impractical assumptions,
our method utilizes generic physical knowledge of power systems.
We design a neural network architecture with the capability of multi-task learning and set the training goal to minimize both voltage prediction errors and power reconstruction errors (i.e., power mismatches\cite{baghaee2017three}).
Our architecture is inspired by supervised auto-encoders\cite{gogna2016semi,le2018supervised} and models the PF solver by minimizing voltage prediction errors,
with an auxiliary task to rebuild PF models from historical measurement data by minimizing power reconstruction errors.
The auxiliary task encodes the basic physical knowledge of reducing power mismatches and acts as a regularizer to alleviate overfitting and improve generalization performance.
In addition, we propose variants of the regularizer, to facilitate the integration of other types of physical knowledge such as the structural property of AC power flow equations\cite{yu2017mapping} and the topology of power systems\cite{zamzam2019physics}.

We evaluate our physics-guided neural networks on the standard IEEE 57 and 118 bus systems.
The simulation results show that our PF solvers achieve an order of magnitude higher accuracy, compared to previous data-driven methods.
Meanwhile, our solvers have better performance on test samples that deviate from training data.
All these results show that by incorporating generic physical knowledge of power systems, our neural networks are regularized and guided to focus on physically consistent solutions, thus have better performance and generalizability.
Furthermore, we demonstrate that the weight matrices of our regularizer variants, which integrates the structural property of PF equations and the topology of power systems, share similar patterns with bus admittance matrices.
This shows not only the effectiveness of physical knowledge integration but also the potential of using our auxiliary task for parameter estimation and PF model rebuilding.

The contributions of this paper can be summarized as follows:
\begin{itemize}
    \item We propose the first physics-guided neural network for PF analysis given the fact that system parameters and PF models can be inaccurate or unavailable.
    \item We quantitatively evaluate and compare all data-driven PF solvers and show that our methods consistently achieve better accuracy and generalizability.
    \item We demonstrate that our two regularizer variants embody physical knowledge in their weight matrices and elaborate corresponding use cases.
    
\end{itemize}

\section{Power Flow Analysis Problem Formulation}
\subsection{Model-based Methods for Power Flow Analysis}
In existing literature, the power flow problem are formulated in two different coordinate systems.

First, the basic active power flow (ACPF) problem can be formulated as a set of nonlinear trigonometric equations in polar coordinate, which represent Kirchhoff's laws \cite{andersson2008modelling}:
\begin{equation}
    \boldsymbol{p}_i = \sum\limits_{k=1}^N\boldsymbol{V}_i\boldsymbol{V}_k(\boldsymbol{G}_{ik}cos\boldsymbol{\theta}_{ik}+\boldsymbol{B}_{ik}sin\boldsymbol{\theta}_{ik})
    \label{eq:ppolar}
\end{equation}
\begin{equation}
    \boldsymbol{q}_i = \sum\limits_{k=1}^N\boldsymbol{V}_i\boldsymbol{V}_k(\boldsymbol{G}_{ik}sin\boldsymbol{\theta}_{ik}-\boldsymbol{B}_{ik}cos\boldsymbol{\theta}_{ik})
    \label{eq:qpolar}
\end{equation}
where $\boldsymbol{p}_i$ and $\boldsymbol{q}_i$ are the real and reactive power injections at bus $i$, $\boldsymbol{G}_{ik}$ and $\boldsymbol{B}_{ik}$ are the real and imaginary parts of the $(i,k)-$th element in the bus admittance matrix respectively, $\boldsymbol{V}_i$ is the voltage magnitude, $\boldsymbol{\theta}_{ik}$ is the voltage phase angle difference between bus $i$ and bus $k$, and $N$ is the total bus number in a power system.

In the basic formulation of the PF problem, three types of buses are defined. 
PQ buses are load buses, the real and reactive power injections of which are specified, but the voltage magnitudes and angles are unknown. 
PV buses are generation buses, the real power injections and voltage magnitudes of which are specified, but the reactive power injections and voltage angles are unknown. 
V$\theta$ bus is the reference bus, where the voltage magnitude and angle are specified, but the real and reactive power injections are unknown. 
For a power system consisting of the three bus types, a set of PF equations
with the same number of equations and unknowns are obtained:
\begin{equation}
    \boldsymbol{g}([\boldsymbol{V}; \boldsymbol{\theta}]) =  \begin{pmatrix} 
      p_1([\boldsymbol{V}; \boldsymbol{\theta}])-p_1^s \\ 
      \vdots \\ 
      p_m([\boldsymbol{V}; \boldsymbol{\theta}])-p_m^s  \\
      q_1([\boldsymbol{V}; \boldsymbol{\theta}])-q_1^s \\ 
      \vdots \\ 
      q_n([\boldsymbol{V}; \boldsymbol{\theta}])-q_n^s 
   \end{pmatrix}=\boldsymbol{0}
   \label{eq:newton}
\end{equation}
where $p_k([\boldsymbol{V}; \boldsymbol{\theta}])$ and $q_k([\boldsymbol{V}; \boldsymbol{\theta}])$ are ACPF equations defined by Equation \ref{eq:ppolar} and \ref{eq:qpolar}, $p_k^s$ and $q_k^s$ are known real and reactive power injections into PQ and PV buses. 
To fully specify the voltages of all buses, numerical methods are conventionally used to iteratively solve Equation \ref{eq:newton}.

Another formulation of the PF problem is to represent ACPF equations in rectangular coordinates.
Let $\boldsymbol{\mu}_i = 
\boldsymbol{V}_icos\boldsymbol{\theta}_i$ and $\boldsymbol{\omega}_i = \boldsymbol{V}_isin\boldsymbol{\theta}_i$ be the real and imaginary components of a voltage phasor, ACPF equations become,
\begin{equation}
    \boldsymbol{p}_i = \sum\limits_{k=1}^N\boldsymbol{G}_{ik}(\boldsymbol{\mu}_i\boldsymbol{\mu}_k + \boldsymbol{\omega}_i\boldsymbol{\omega}_k) + \boldsymbol{B}_{ik}(
    \boldsymbol{\omega}_i\boldsymbol{\mu}_k - \boldsymbol{\mu}_i\boldsymbol{\omega}_k)
    \label{eq:prect}
\end{equation}
\begin{equation}
    \boldsymbol{q}_i = \sum\limits_{k=1}^N\boldsymbol{G}_{ik}(\boldsymbol{\omega}_i\boldsymbol{\mu}_k - \boldsymbol{\mu}_i\boldsymbol{\omega}_k) - \boldsymbol{B}_{ik}(\boldsymbol{\mu}_i\boldsymbol{\mu}_k + \boldsymbol{\omega}_i\boldsymbol{\omega}_k)
    \label{eq:qrect}
\end{equation}
The power injection $\boldsymbol{p}_i$ and $\boldsymbol{q}_i$ are second-order polynomial functions of the voltage vector $[\boldsymbol{\mu}, \boldsymbol{\omega}]$. 
%

\subsection{From Model-based Methods to Data-driven Methods}
Numerical methods typically require accurate PF equations.
Therefore, numerical PF solvers are subject to the inaccuracy and unavailability of PF models.
The issues of modeling the PF problem usually come from the missing knowledge of system parameters and control rules. 
Especially, in distribution grids, it is hard for substation engineers to construct admittance matrices due to the inaccurate or missing information of line parameters \cite{yu2017patopa}.
Modeling DERs is also challenging since DERs are usually independently owned, and manufacturers do not readily provide model data and control policies \cite{guttromson2002modeling}.

Intuitively, numerical PF solvers are expected to be feasible again by employing data-driven approaches~\cite{yuan2016inverse, yu2017patopa, chen2016measurement, chen2013measurement} to rediscover system parameters to calibrate or rebuild PF models. 
However, errors are accumulated in the parameter rediscovery and numerical approximation processes\cite{liu2018data}. 
Numerical solutions are sensitive to differences in system parameters. 
A small variance in system parameters can induce big disturbance on the system operation setting, potentially causing errors to ``blow-up". 
In addition to error accumulation, convergence is a problem when DERs are incorporated into power grids, because the power electronic interface provides some degrees of freedom in the models of DER units~\cite{nikkhajoei2006steady, arulampalam2004control}. Therefore, a new power-flow solver is required to solve a set of non-linear equations with an unequal number of equations and unknowns.

\subsection{Data-Driven Power Flow Solvers}
Data-driven power flow solvers produce PF solutions by inferring the mapping rules between the observed system inputs and outputs based on historical system operation data. 
To obtain voltage values of all buses, data-driven methods use machine learning algorithms to solve Equation \ref{eq:newton} by learning the mapping rules from specified variables to bus voltages:
\begin{equation}
    [\boldsymbol{\mu}; \boldsymbol{\omega}] = f([\boldsymbol{P}_L; \boldsymbol{P}_G; \boldsymbol{Q}_L; \boldsymbol{V}_G; V_R; \theta_R])
    \label{eq:inverse_rec}
\end{equation}
where, according to the bus type definition, the specified variables or power-flow inputs are power injections $\boldsymbol{P}_L$, $\boldsymbol{Q}_L$ on load buses, real power injections and voltage magnitudes $\boldsymbol{P}_G$, $\boldsymbol{V}_G$ on generation buses, and voltages $V_R$, $\theta_R$ on the reference bus.

The challenges of data-driven PF solvers come from two main aspects: to avoid overfitting and to improve generalizability.
Since the PF solutions, i.e., Equation \ref{eq:inverse_rec}, do not have closed-form expressions, there is no guide to select a proper machine learning model family.
Besides, as measurement data may not be various to reflect all aspects of power systems, data-driven methods may have generalization issues.
Incorporating physics knowledge of power systems into data-driven models is a promising way to reduce overfitting and improve generalizability~\cite{baker2019workshop}
and several neural-network PF solvers~\cite{muller2010artificial,baghaee2017three} have explored to this path.
M{\"u}ller et al~\cite{muller2010artificial} incorporates physics by treating system parameters as known inputs (e.g., bus admittance matrix).
Similarly, Baghaee et al~\cite{baghaee2017three} assumes the availability of accurate PF models, and
considers physics by repeatedly retraining RBFNN until the power mismatch of the RBFNN outputs reaches a predefined tolerance. 
Unfortunately, both of these methods cannot consider physics if system parameters and the PF models are inaccurate or unavailable. 
How to incorporate generic physics laws into data-driven methods without reliance on accurate system models still remains an open problem.

\section{Physics-Guided Deep Neural Network for Power Flow Analysis}
In this section, we describe how we add regularization terms to a ``black-box'' multilayer perceptron neural network (MLPNN) to solve the PF problem under the constraints of generic physical knowledge of power systems.

\subsection{Data-Driven Power Flow Solver with Regularizer}
\begin{figure}[h]
\vspace{-4mm}
\includegraphics[width=.4\textwidth]{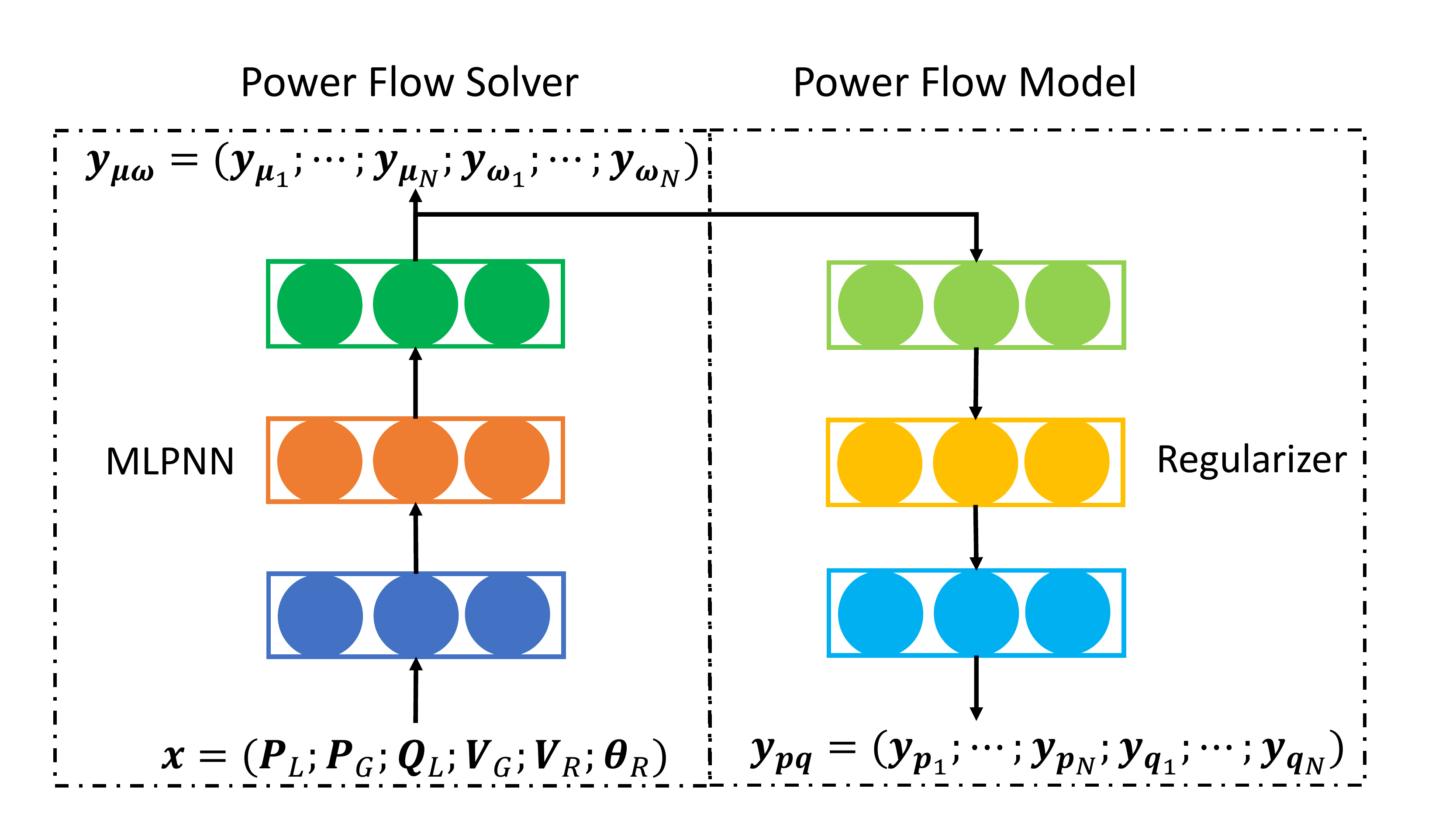}
\centering
\vspace{-4mm}
\caption{Neural network power flow solver with regularization.}
\vspace{-2mm}
\label{fig:architecture}
\end{figure}

In spite of the universal function approximation theorem of MLPNN, multiple relations in the hypothesis space of MLPNN can map power-flow inputs to bus voltages.
To guide the hypothesis search procedure, we introduce regularizers through multi-task learning.
The architecture of our physics-guided data-driven PF solver with regularization is illustrated in Fig. \ref{fig:architecture}.
It is inspired by supervised AutoEncoder~\cite{le2018supervised} and consists of two parts respectively focusing on the inverse and forward problems: power flow solving and power flow modeling.

The ``encoder" part is a black-box MLPNN focusing on solving the PF problem by learning the mapping from power-flow inputs $\boldsymbol{x}$ to  bus voltages $[\boldsymbol{\mu};\boldsymbol{\omega}]$. 
Assume the MLPNN ``encoder'' has 3 layers, its outputs are
\begin{align}
\begin{split}
    \boldsymbol{y_{\mu\omega}}{}&=f_{en}(\boldsymbol{x}; \boldsymbol{W}_{en}, \boldsymbol{b}_{en})\\
    {}&=
    \sigma_{3}(
    \boldsymbol{W}_{3}\sigma_{2}(
    \boldsymbol{W}_{2}\sigma_{1}(
    \boldsymbol{W}_{1}\boldsymbol{x}
    +\boldsymbol{b}_{1})
    +\boldsymbol{b}_{2})
    +\boldsymbol{b}_{3})
\end{split}
\end{align}
where $\sigma_{t}()$ is a point-wise activation function at the $t$-th layer, $\mathbf{W}_{t}$ and $\mathbf{b}_{t}$ are respectively the weight matrix and bias of the $t$-th layer. 

The ``decoder" part is a neural network carrying out the auxiliary task, which is rebuilding PF models mapping bus voltages to power injections. 
We will discuss the structure of the ``decoder'' neural network later. 
Here let's denote the outputs of the ``decoder'' as
\begin{equation}
    \boldsymbol{y_{pq}}=f_{de}(\boldsymbol{y_{\mu\omega}}; \boldsymbol{W}_{de}, \boldsymbol{b}_{de})
\end{equation}

The loss function  is a weighted sum of voltage prediction losses produced by the ``encoder'' and power injection reconstruction losses produced by the ``decoder''.
Let $\boldsymbol{\Delta\mu\omega}=\boldsymbol{[\mu;\omega]-\boldsymbol{y_{\mu\omega}}}$, $\boldsymbol{\Delta pq}=\boldsymbol{[p;q]-\boldsymbol{y_{pq}}}$, the loss for a single training sample is 
\begin{equation}
\begin{split}
    \ell =  \alpha_{sup} \ell(\boldsymbol{\Delta\mu\omega})
    + \alpha_{unsup}\ell(\boldsymbol{\Delta pq})
\end{split}
    \label{eq:objective}
\end{equation}
where typically $\ell()$  is the squared error function.

A variety of works \cite{maurer2006bounds, maurer2016benefit, liu2016algorithm, le2018supervised} have theoretically and empirically demonstrated that auxiliary task or multi-task learning serves as a regularizer to constrain the solution and improve generalization performance.
When training our neural network, the errors at the voltage prediction layer is
\begin{equation}
    \frac{\partial \ell}{\partial \boldsymbol{y_{\mu\omega}}}= -\alpha_{sup}\boldsymbol{\Delta\mu\omega}+\alpha_{unsup}\boldsymbol{\Delta pq}\frac{\partial\boldsymbol{\Delta pq}}{\partial \boldsymbol{y_{\mu\omega}}}
    \label{eq: errors_multitask}
\end{equation}
where $\frac{\partial\boldsymbol{\Delta pq}}{\partial \boldsymbol{y_{\mu\omega}}}$ is similar to the PF Jacobian matrix $\boldsymbol{J}$.
The first term $\boldsymbol{\Delta\mu\omega}$ directs learning the MLPNN ``encoder'' towards parameters that are effective for minimizing bus voltage prediction errors.
This is desired and the main training goal.
However, solely training an MLPNN according to the supervised losses is an under-constrained problem, and will find parameters that over-fit the data and do not generalize well.
With the guide of the second term $\boldsymbol{\Delta pq}\frac{\partial\boldsymbol{\Delta pq}}{\partial \boldsymbol{y_{\mu\omega}}}$, among similarly effective parameters that produce good voltage estimation, the learning algorithm will prefer the one that generates small power mismatches. 
In this way, the combination of the two types of losses is supposed to generate a more robust and physically consistent data-driven PF solver.

\subsection{Encode Physical Knowledge in Regularizer}
In this section, we describe how we design different ``decoders'' to rebuild PF models according to different prior physical knowledge of power systems.

\paragraph{MLPNN Regularizer}
To calculate power reconstruction errors without PF models, the intuitive method is to employ MLPNNs to learn any functions to approximate PF equations.
Since active and reactive power injections are different nonlinear functions of bus voltages, we decouple their mapping rule learning process and utilize two MLPNNs with a shared input layer.
Given the same inputs, i.e. bus voltages estimation $\boldsymbol{y_{\mu\omega}}$ of the ``encoder'', the output layers of these two MLPNNs predict active and reactive power injections respectively:
\begin{equation}
    \boldsymbol{y_p} =
    \sigma_{6}(\boldsymbol{W}_{6}
    \sigma_{5}(\boldsymbol{W}_{5}
    \sigma_{4}(\boldsymbol{W}_{4}\boldsymbol{y_{\mu\omega}} 
    + \boldsymbol{b}_{4})
    + \boldsymbol{b}_{5})
    + \boldsymbol{b}_{6})
\end{equation}
\begin{equation}
    \boldsymbol{y_q} =
    \sigma_{9}(\boldsymbol{W}_{9}
    \sigma_{8}(\boldsymbol{W}_{8}
    \sigma_{7}(\boldsymbol{W}_{7}\boldsymbol{y_{\mu\omega}} 
    + \boldsymbol{b}_{7})
    + \boldsymbol{b}_{8})
    + \boldsymbol{b}_{9})
\end{equation}
where we assume both MLPNNs have 3 layers. We call the neural network with the MLP ``encoder'' and the MLP ``decoder'' as ``MLP+MLP'' PF solver.
%
%

\begin{figure}[h]
\vspace{-4mm}
\includegraphics[width=.4\textwidth]{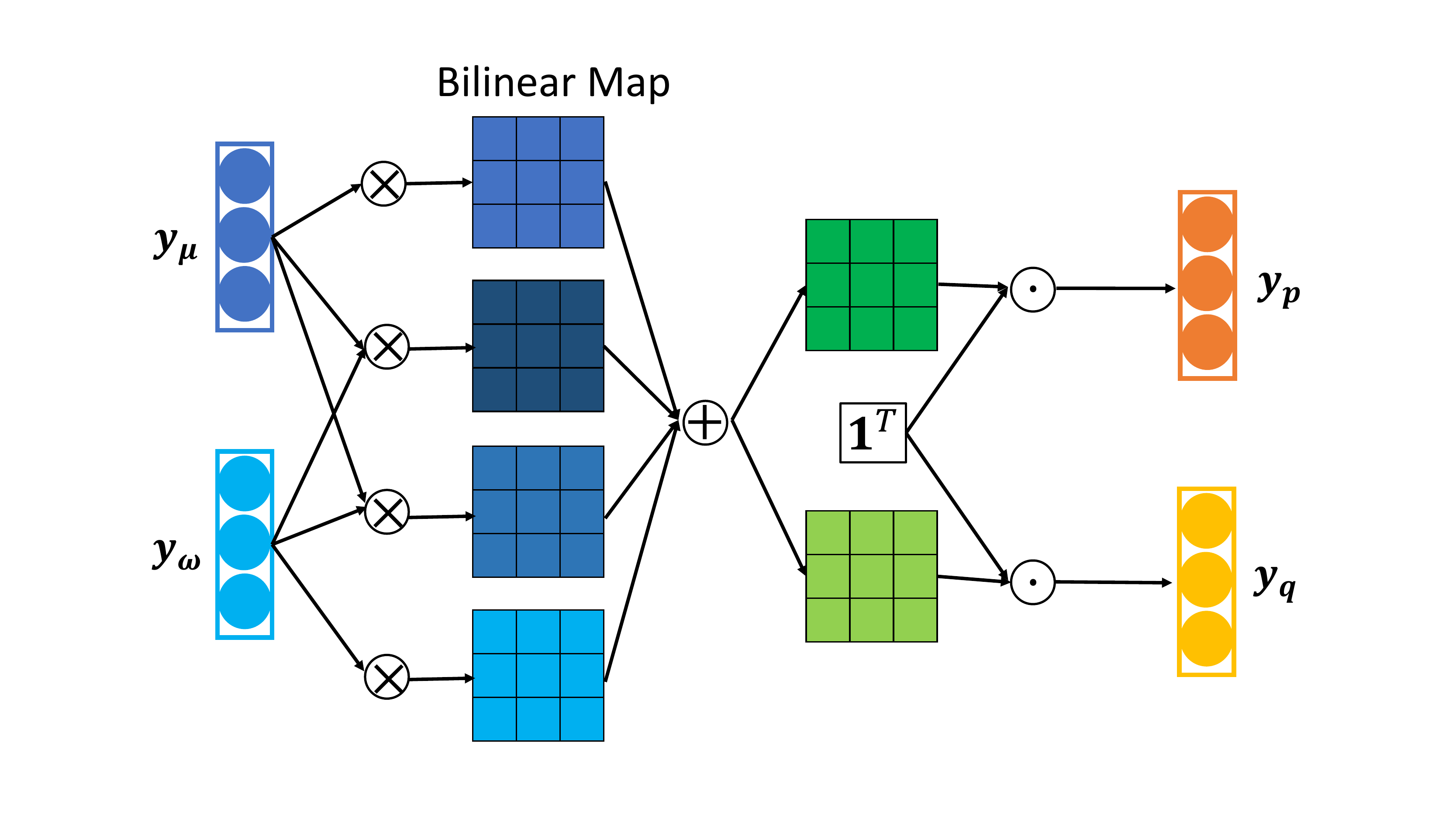}
\centering
\vspace{-4mm}
\caption{Bilinear neural network to rebuild power flow models.}
\vspace{-2mm}
\label{fig:bnn_architecture}
\end{figure}

\paragraph{BNN Regularizer}
Different from the MLPNN regularizer, we design a ``decoder'' by utilizing the structural properties of ACPF equations to approximate PF models and incorporate power reconstruction errors.
As shown in Equation \ref{eq:prect} and \ref{eq:qrect}, a power injection is basically a linear combination of the 2-th Kronecker power of bus voltages in the rectangular coordinate. 
This bilinear function formulation is indicated by Kirchhoff's laws, except that the function parameters, i.e. bus admittance matrix $\boldsymbol{G}$ and $\boldsymbol{B}$, may be unavailable or inaccurate due to system dynamics, uncertainties, and erroneous measurements.
To utilize this prior knowledge of Kirchhoff's laws which indicate the structure of PF models, we design a Bilinear Neural Network (BNN), depicted in Fig. \ref{fig:bnn_architecture}.

To learn the bilinear mapping rules from bus voltages to power injections, BNN first builds bilinear maps based on the outer products of the given voltage vectors $\boldsymbol{y_\mu}=[y_{\mu_1};y_{\mu_2};\dots;y_{\mu_n}]$ and $\boldsymbol{y_\omega}=[y_{\omega_1};y_{\omega_2};\dots;y_{\omega_n}]$.
Then, these bilinear maps are weighted summed by weight matrices $\boldsymbol{W_G}$ and $\boldsymbol{W_B}$, which are the weight matrices of BNN and are related to bus admittance matrices.
\begin{equation}
    \boldsymbol{S}_{1} = (\boldsymbol{y_\mu}\boldsymbol{y}_{\boldsymbol{\mu}}^T 
    + \boldsymbol{y_\omega}\boldsymbol{y}_{\boldsymbol{\omega}}^T) 
    \circ \boldsymbol{W_G} 
    + (\boldsymbol{y_\omega}\boldsymbol{y}_{\boldsymbol{\mu}}^T 
    - \boldsymbol{y_\mu}\boldsymbol{y}_{\boldsymbol{\omega}}^T) \circ \boldsymbol{W_B}
    \label{eq:bilinear_product1}
\end{equation}
\begin{equation}
    \boldsymbol{S}_{2} = (\boldsymbol{y_\omega}\boldsymbol{y}_{\boldsymbol{\mu}}^T 
    - \boldsymbol{y_\mu}\boldsymbol{y}_{\boldsymbol{\omega}}^T) \circ \boldsymbol{W_G} 
    - (\boldsymbol{y_\mu}\boldsymbol{y}_{\boldsymbol{\mu}}^T + \boldsymbol{y_\omega}\boldsymbol{y}_{\boldsymbol{\omega}}^T) \circ \boldsymbol{W_B}
    \label{eq:bilinear_product2}
\end{equation}
By multiplying an all-one vector $\boldsymbol{1}$ to the bilinear maps $\boldsymbol{S}_{1}$ and $\boldsymbol{S}_{2}$, we flatten the bilinear maps and further linearly combine the bilinear terms to predict power injections:
\begin{equation}
    \boldsymbol{y}_p = \boldsymbol{S}_{1}\boldsymbol{1}^T + \boldsymbol{b_p}
\end{equation}
\begin{equation}
    \boldsymbol{y}_q = \boldsymbol{S}_{2}\boldsymbol{1}^T + \boldsymbol{b_q}
\end{equation}
where bias terms $\mathbf{b_p}, \mathbf{b_q}$ are added to absorb noise in the input voltage vectors and enhance the regression capability.
We call the neural network with the MLP ``encoder'' and the BNN ``decoder'' as ``MLP+BNN'' PF solver.

\paragraph{TPBNN Regularizer}
Built upon the BNN architecture, we can further encode topology information into the ``decoder'' module if system topology is available.
When bus $i$ and bus $j$ are not directly connected, their corresponding system parameters $\boldsymbol{G}_{ij}$, $\boldsymbol{G}_{ji}$, $\boldsymbol{B}_{ij}$, and $\boldsymbol{B}_{ji}$ are zero. 
Based on this fact, the weights of monomials $\boldsymbol{y_\mu}_i\boldsymbol{y_\mu}_j$, $\boldsymbol{y_\mu}_i\boldsymbol{y_\omega}_j$, $\boldsymbol{y_\omega}_i\boldsymbol{y_\mu}_j$, and $\boldsymbol{y_\omega}_i\boldsymbol{y_\omega}_j$ in BNN can be zeroed out. 
Therefore, in Equation \ref{eq:bilinear_product1} and Equation \ref{eq:bilinear_product2}, some parameters in the weight matrices $\boldsymbol{W_G}$ and $\boldsymbol{W_B}$ are redundant.

To reduce the redundancy in the parameter space and improve generalization, when the topology of a power grid is available, we prune the parameters of BNN by employing the adjacent matrix as a hard attention mask.
Let $\boldsymbol{A}$ be the adjacent matrix of a power system.
The Hadamard product of bus admittance matrix and adjacent matrix is bus admittance matrix itself, i.e. $\boldsymbol{G} \circ \boldsymbol{A} = \boldsymbol{G}$, $\boldsymbol{B} \circ \boldsymbol{A} = \boldsymbol{B}$.
As the role of weight matrices $\boldsymbol{W_G}$ and $\boldsymbol{W_B}$ is equivalent to the role of bus admittance matrices, we expect that $\boldsymbol{W_G} \circ \boldsymbol{A} = \boldsymbol{W_G}$, $\boldsymbol{W_B} \circ \boldsymbol{A} = \boldsymbol{W_B}$.
Therefore, by augmenting topology knowledge to BNN, the power injections are calculated by
\begin{equation}
    \boldsymbol{y}_p'= (\boldsymbol{S}_{1} \circ \boldsymbol{A}) \boldsymbol{1}^T + \boldsymbol{b_p}
\end{equation}
\begin{equation}
    \boldsymbol{y}_q' = (\boldsymbol{S}_{2} \circ \boldsymbol{A})\boldsymbol{1}^T + \boldsymbol{b_q}
\end{equation}
In the process of backpropagation, compared to the corresponding BNN, the gradients of the $(i,j)$-th element of the weight matrices $\boldsymbol{W_G}$ and $\boldsymbol{W_B}$ are
\begin{equation}
    \frac{\partial \ell(\boldsymbol{y}_p', \boldsymbol{p}) + \ell(\boldsymbol{y}_q', \boldsymbol{q})}{\partial \boldsymbol{W}_{\boldsymbol{G}ij}} = \boldsymbol{A}_{ij}\frac{\partial \ell(\boldsymbol{y}_p, \boldsymbol{p}) + \ell(\boldsymbol{y}_q, \boldsymbol{q})}{\partial \boldsymbol{W}_{\boldsymbol{G}ij}}
\end{equation}
\begin{equation}
    \frac{\partial \ell(\boldsymbol{y}_p', \boldsymbol{p}) + \ell(\boldsymbol{y}_q', \boldsymbol{q})}{\partial \boldsymbol{W}_{\boldsymbol{B}ij}} = \boldsymbol{A}_{ij}\frac{\partial \ell(\boldsymbol{y}_p, \boldsymbol{p}) + \ell(\boldsymbol{y}_q, \boldsymbol{q})}{\partial \boldsymbol{W}_{\boldsymbol{B}ij}}
\end{equation}
When the bus $i$ and bus $j$ are not neighbors, the corresponding $\mathbf{A}_{ij}$ is zero and the gradients of $\mathbf{W}_{\mathbf{G}ij}$ and $\mathbf{W}_{\boldsymbol{B}ij}$ are zeros.
By initializing $\mathbf{W}_{\mathbf{G}ij}$ and $\mathbf{W}_{\boldsymbol{B}ij}$ as zeros, the weight matrices $\mathbf{W_G}$ and $\mathbf{W_B}$ will keep the same sparsity pattern as admittance matrices.
In this way, we encode the topology information into the BNN by using the Hadamard multiplication of the adjacent matrix and the weight matrices.

We call the BNN of topology information as topology-pruned BNN (TPBNN), and the neural network with the MLP ``encoder'' and the TPBNN ``decoder'' as ``MLP+TPBNN'' PF solver.

\section{Experimental Results}
\subsection{Experiment Setup}
We test our physics-guided NNs for PF analysis on the IEEE 57 and 118 bus systems. 
To simulate the power system operation measurement data, we use the real-world load data from GEFCom 2012~\cite{hong2014global} as load consumption, where load series were subsampled and scaled to match and balance the system generation capacity.
Then, the MATPOWER~\cite{zimmerman2010matpower} is employed to obtain the associated voltage magnitudes and phase angles at each bus. 
Unless otherwise stated, we use the first 60\% of samples as training data, the next 10\% of samples as validation data, and the last 30\% of samples as testing data. 
Besides, Gaussian noise with a 1\% relative standard deviation is added to the training data because measurement noise generally exists in practical settings. 
While we evaluate our methods, we assume that the accurate line parameters are unavailable. 
We fit our physics-guided NNs on the training dataset to rebuild and solve PF models with the awareness of physical laws. 
To alleviate randomness in the learning process of neural networks, we train and test all NNs for 10 times and average the results. 
The source code is available online
\footnote{Source code will be made available once published.}.

\subsection{Power Flow Solution Results}
\begin{table*}
    \caption{Power Flow Solver Average Results}
    \centering
    \resizebox{\textwidth}{!}{%
    \begin{tabular}{c|c||c|c|c|c|c|c}
\hline
\multicolumn{2}{c||}{ } & \multicolumn{6}{|c}{Root Mean Squared Error (p.u.)}\\
 \hline
 Case& V &LR~\cite{liu2018data} & SVR~\cite{yu2017mapping} & MLP~\cite{muller2010artificial} & MLP+MLP & MLP+BNN & MLP+TPBNN\\
 \hline
 \multirow{2}{*}{IEEE 57} & $\boldsymbol{\mu}$ & $1.95\times10^{-2}$ &$4.81\times10^{-3}$& $1.06\times10^{-2}\pm3.23\times10^{-4}$ & $3.28\times10^{-3}\pm5.10\times10^{-4}$ & $2.63\times10^{-3}\pm2.16\times10^{-4}$& $\mathbf{{2.41\times10^{-3}\pm2.67\times10^{-4}}}$\\
 & $\boldsymbol{\omega}$ & $8.69\times10^{-3}$ &$5.43\times10^{-3}$& $9.19\times10^{-3}\pm4.49\times10^{-4}$ & $2.63\times10^{-3}\pm5.39\times10^{-4}$ & $2.32\times10^{-3}\pm1.41\times10^{-4}$ & $\mathbf{1.73\times10^{-3}\pm1.43\times10^{-4}}$ \\
 \hline
 \multirow{2}{*}{IEEE 118} &$\boldsymbol{\mu}$ & $7.92\times10^{-2}$ & $2.98\times10^{-2}$ & $2.32\times10^{-2}\pm3.54\times10^{-4}$ & $7.72\times10^{-3}\pm5.83\times10^{-4}$ & $6.55\times10^{-3}\pm1.94\times10^{-4}$ & $\mathbf{5.79\times10^{-3}\pm1.73\times10^{-4}}$\\
 & $\boldsymbol{\omega}$ & $5.52\times10^{-2}$ & $3.89\times10^{-2}$ & $2.49\times10{-2}\pm1.06\times10^{-3}$ &$7.65\times10^{-3}\pm5.93\times10^{-4}$ & $6.37\times10{-3}\pm2.17\times10^{-4}$ & $\mathbf{5.84\times10^{-3}\pm1.65\times10^{-4}}$\\
 \hline
\end{tabular}
    }
    \vspace{-2mm}
    \label{tab:inverse_meanae}
\end{table*}
\begin{figure*}[!t]
\centering
\vspace{-6mm}
\subfloat[MAPE distribution of $\boldsymbol{\mu}$ estimation on the IEEE 57]{\includegraphics[width=0.24\textwidth]{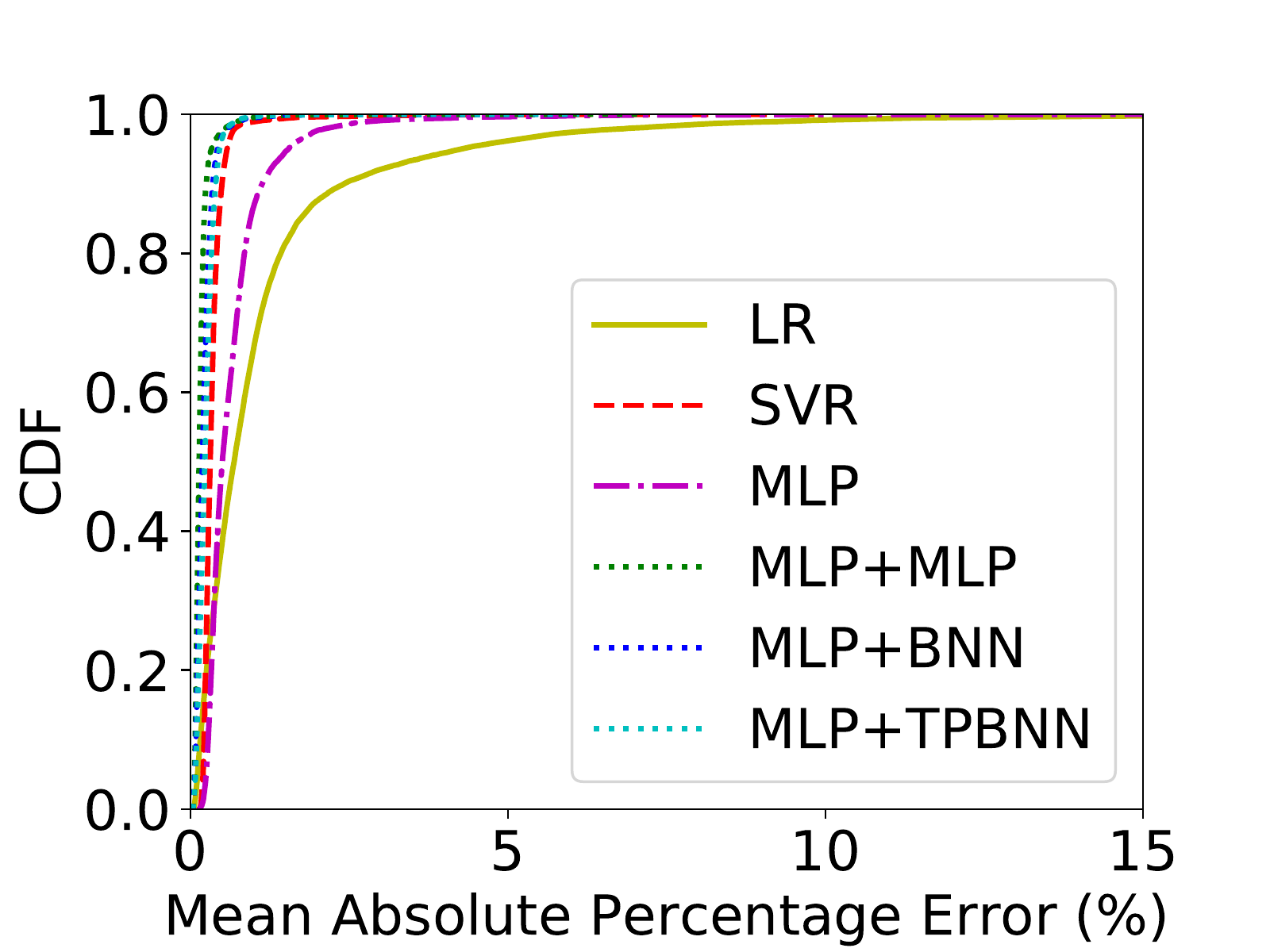}%
\label{fig:57_U_CDF}}
\hfill
\subfloat[MAPE distribution of $\boldsymbol{\omega}$ estimation on the IEEE 57]{\includegraphics[width=0.24\textwidth]{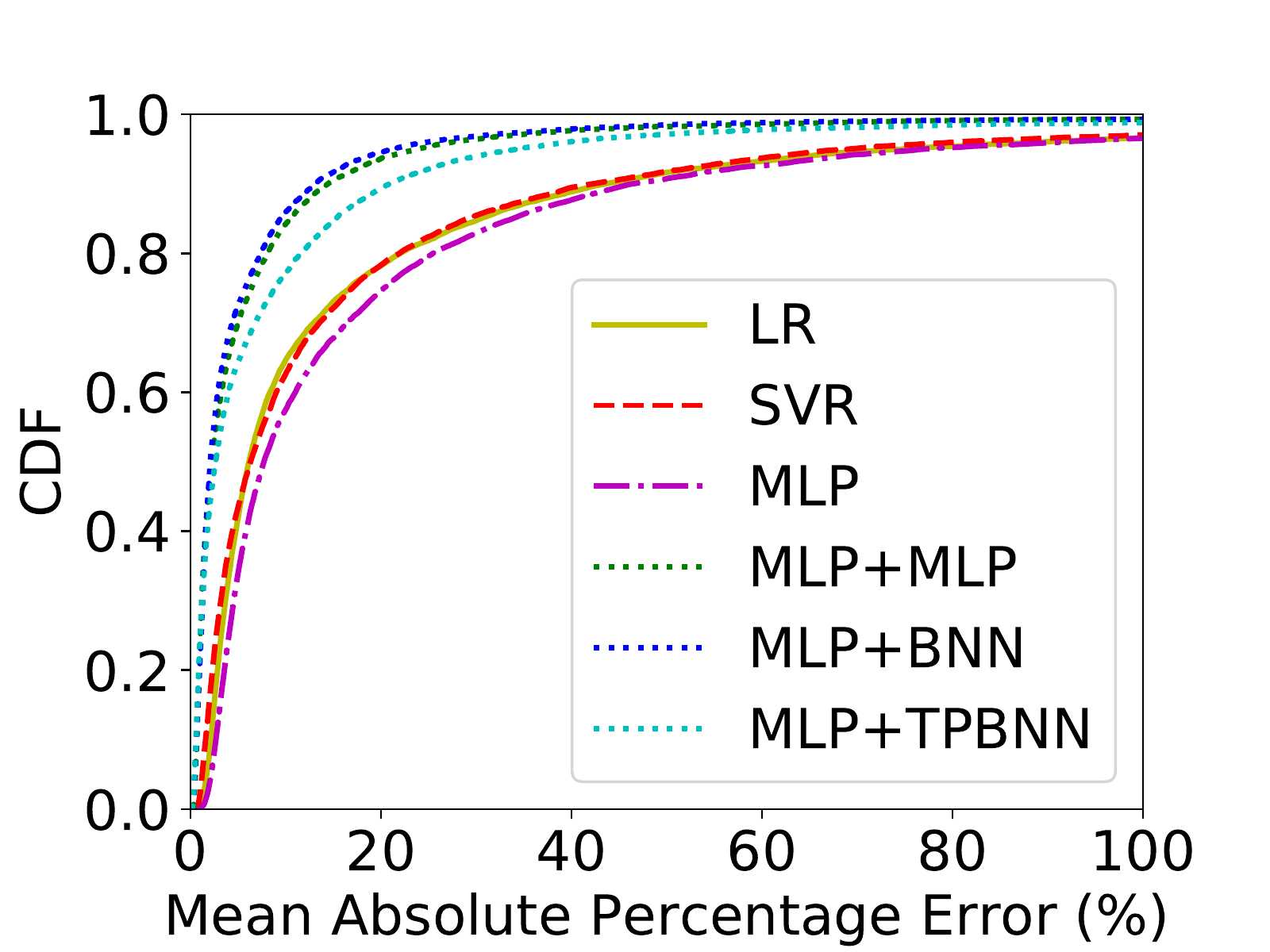}%
\label{fig:57_W_CDF}}
\hfill
\subfloat[MAPE distribution of $\boldsymbol{\mu}$ estimation on the IEEE 118]{\includegraphics[width=0.24\textwidth]{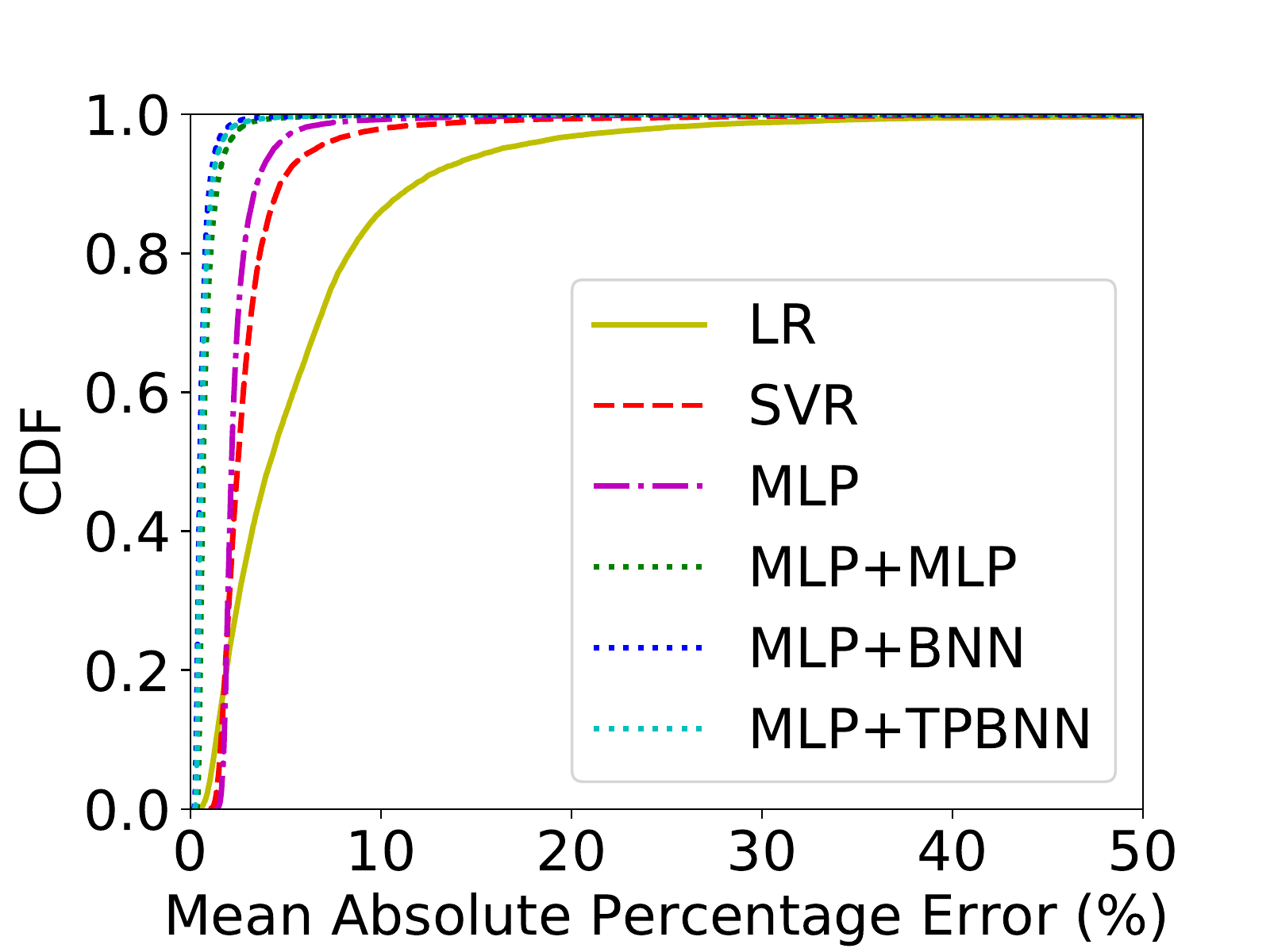}%
\label{fig:118_U_CDF}}
\hfill
\subfloat[MAPE distribution of $\boldsymbol{\omega}$ estimation on the IEEE 118]{\includegraphics[width=0.24\textwidth]{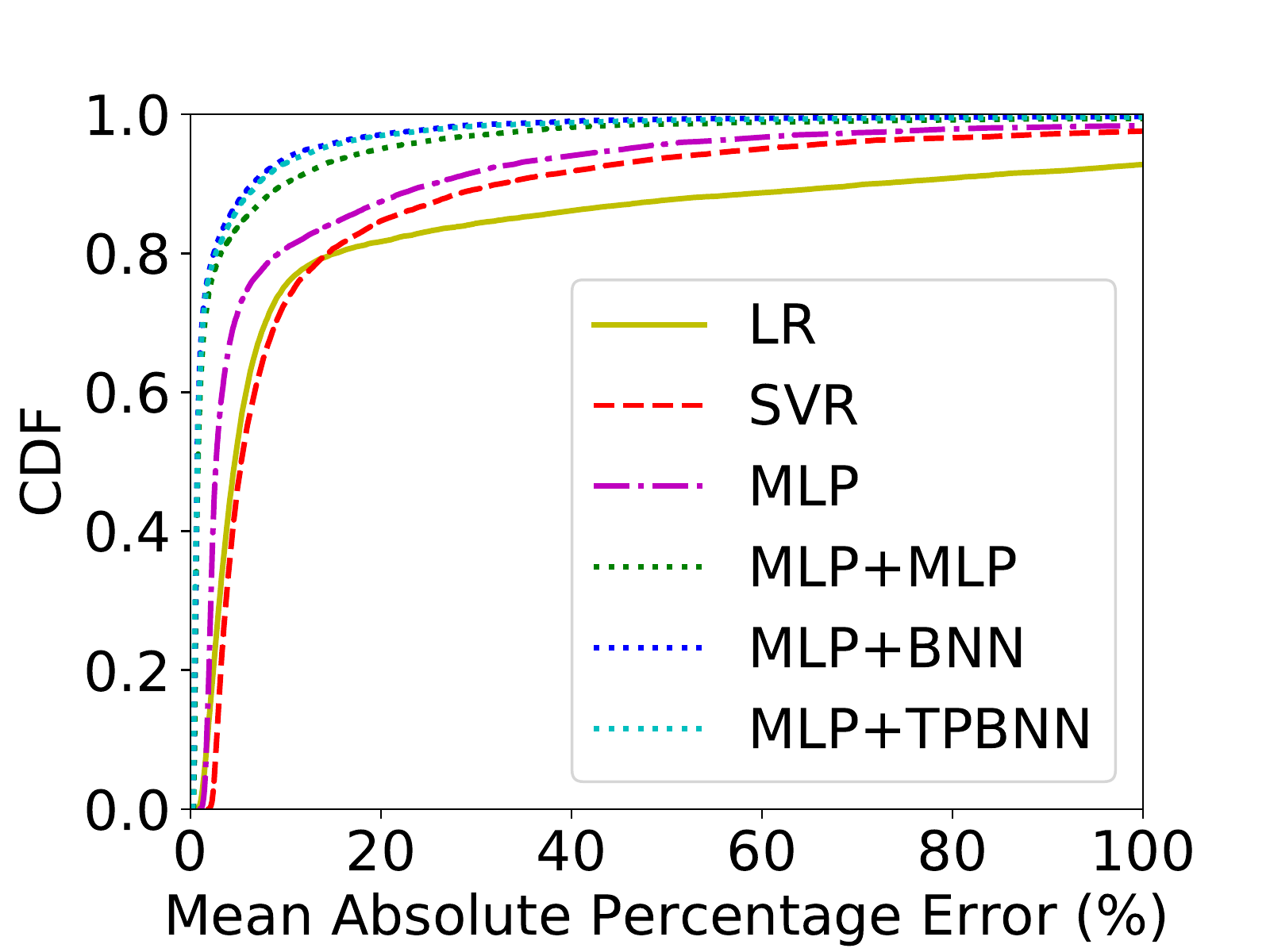}%
\label{fig:118_W_CDF}}
\caption{CDF of mean absolute percentage errors of power flow solutions.}
\vspace{-2mm}
\label{fig:inverse_CDF}
\end{figure*}
%
\begin{figure*}[!t]
\centering
\vspace{-6mm}
\subfloat[Interpolation results on the IEEE 57 case]{\includegraphics[width=0.4\textwidth]{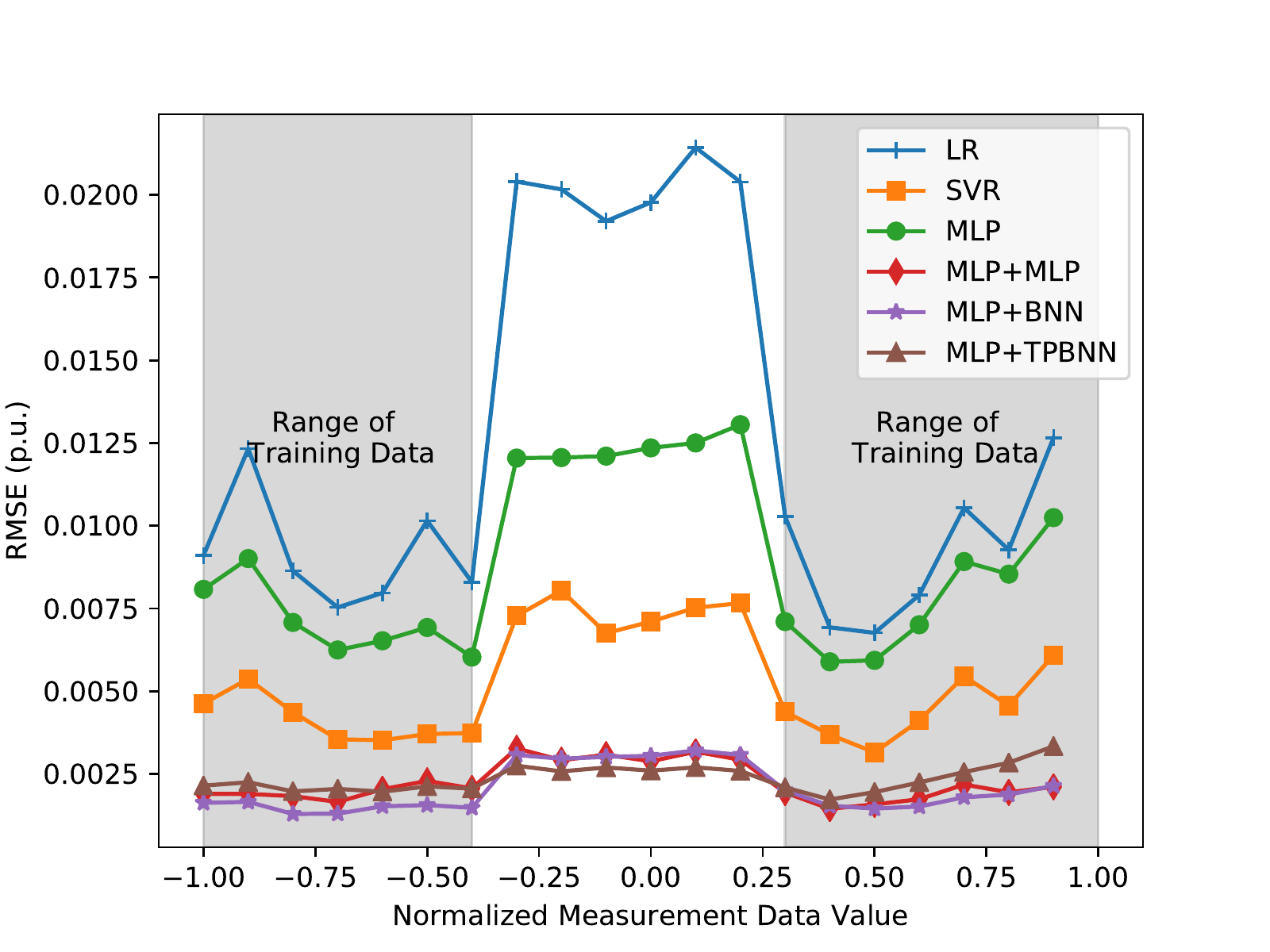}%
\label{fig:57_interpolation}}
\subfloat[Extrapolation results on the IEEE 57 case]{\includegraphics[width=0.4\textwidth]{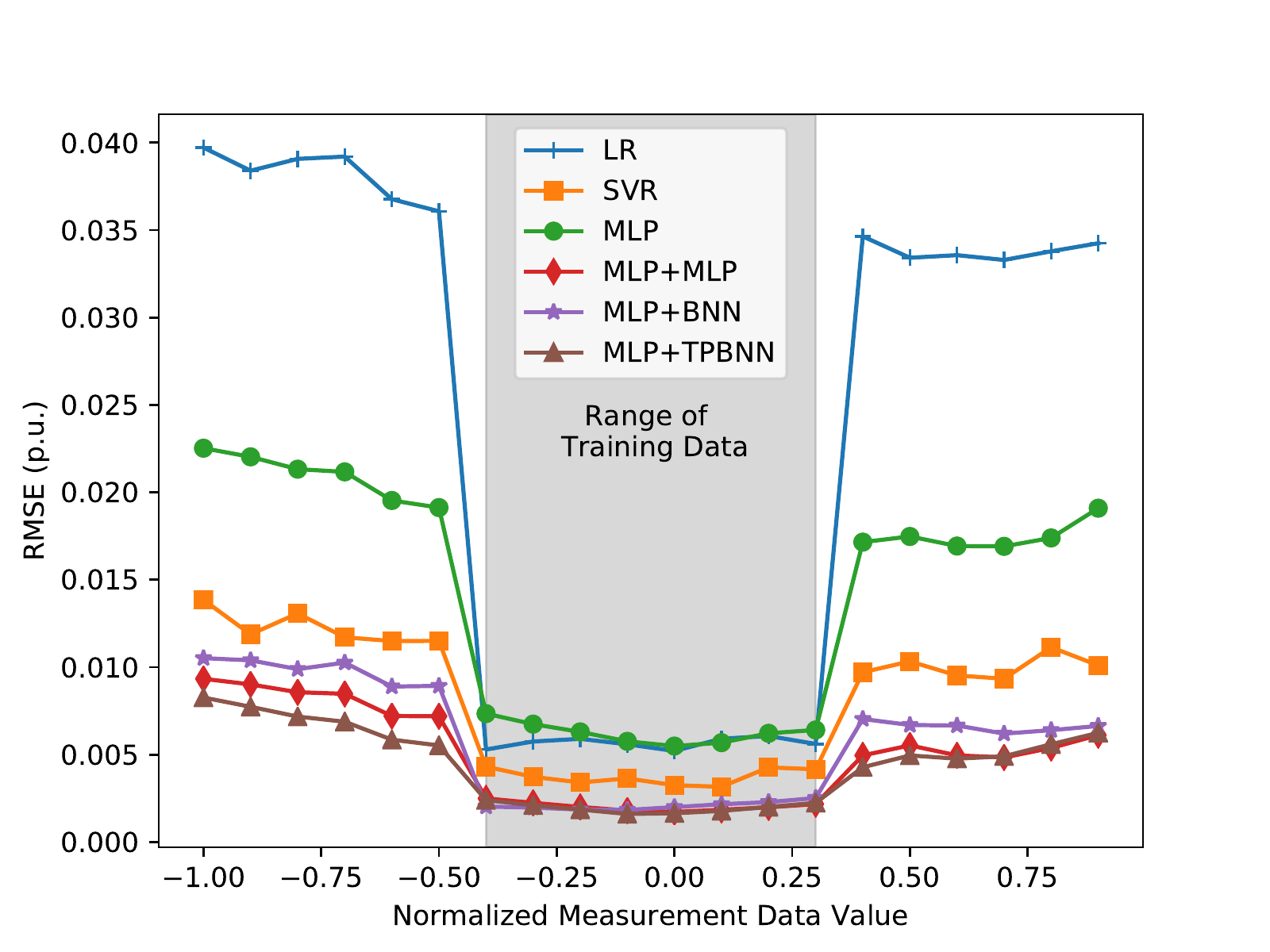}%
\label{fig:57_extrapolation}}
\vspace{-4mm}
\subfloat[Interpolation results on the IEEE 118 case]{\includegraphics[width=0.4\textwidth]{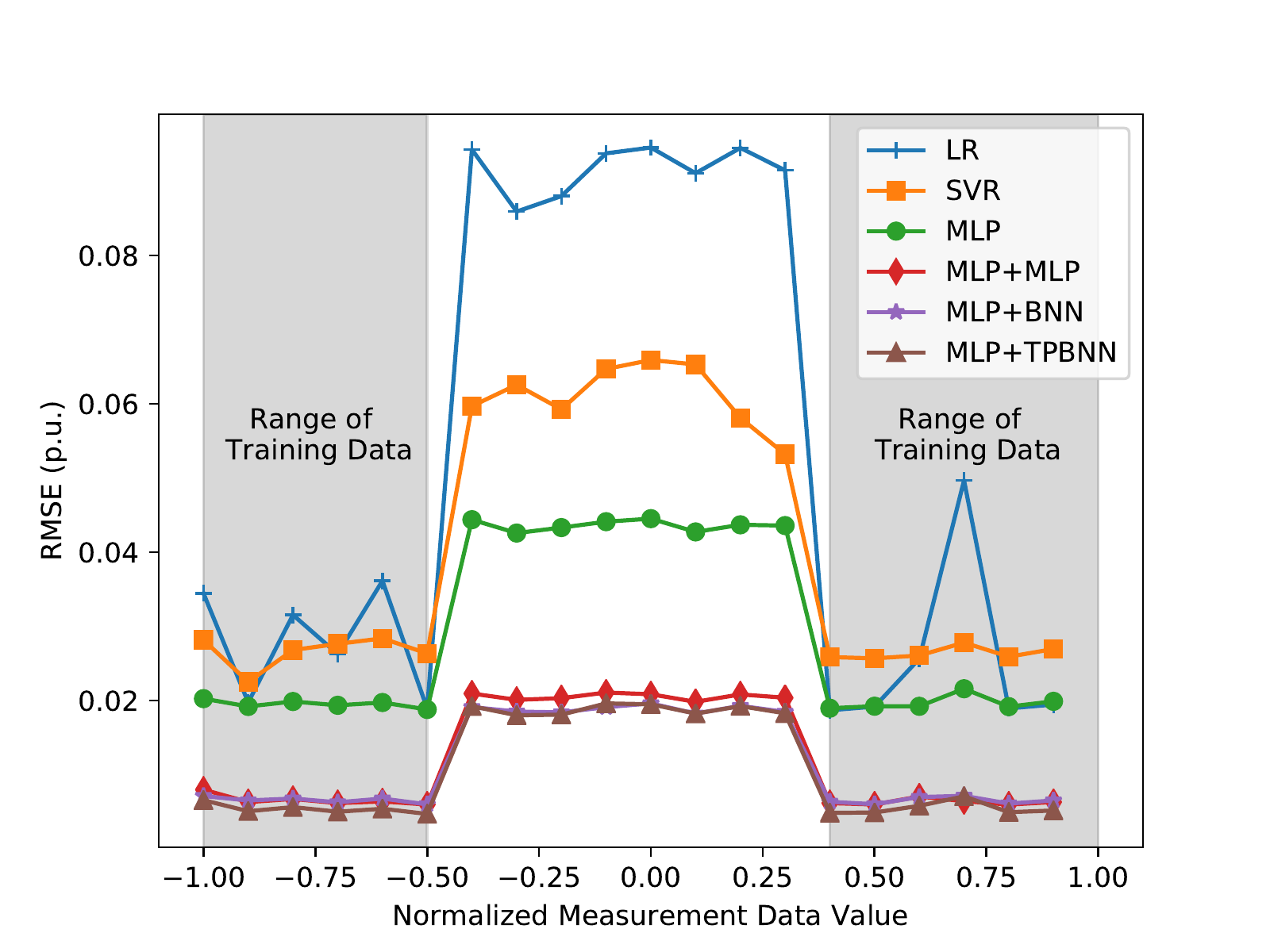}%
\label{fig:118_interpolation}}
\subfloat[Extrapolation results on the IEEE 118 case]{\includegraphics[width=0.4\textwidth]{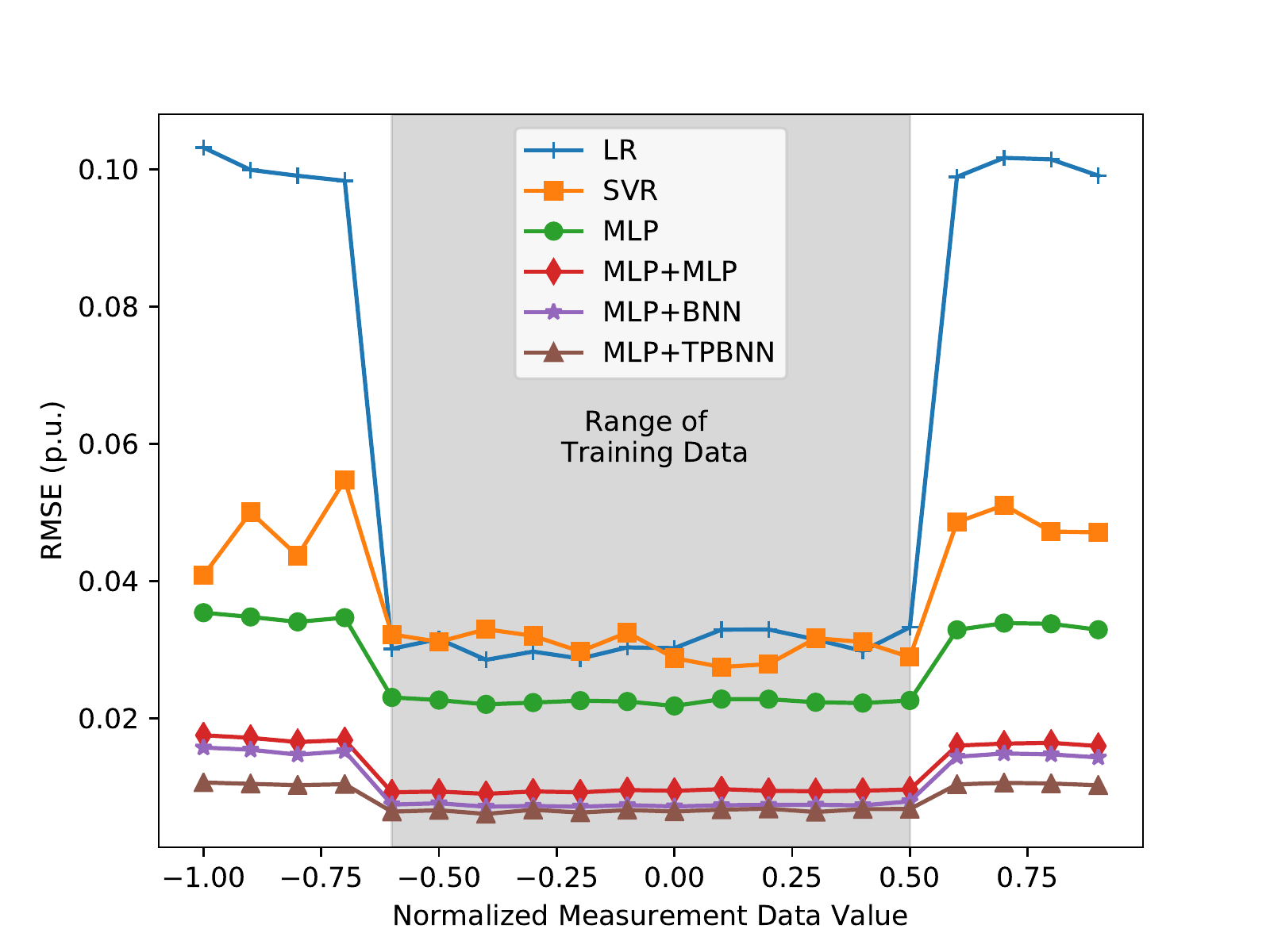}%
\label{fig:118_extrapolation}}
\caption{Distribution of interpolation and extrapolation results.}
\vspace{-4mm}
\label{fig:inter_extra_polation}
\end{figure*}
In this section, we evaluate the effectiveness of the three proposed physics-guided neural network PF solvers.
To compare our methods with existing data-driven power-flow solvers, we modify the LR~\cite{liu2018data}, SVR~\cite{yu2017mapping}, and MLPNN~\cite{muller2010artificial} methods, such that they have the same inputs (i.e. power-flow inputs) and outputs (i.e. bus voltages) with our methods. 
We skip the RBFNN method~\cite{karami2008radial} since RBFNN has similar capability (i.e. universal function approximation) and performance compared to MLPNN.

\subsubsection{Basic Results}
Table \ref{tab:inverse_meanae} shows the average results of different data-driven PF solvers in terms of RMSE (Root Mean Square Error). 
The average voltage errors of the LR, SVR, and MLPNN methods on the two test cases are $6.02\times10^{-2}$, $1.97\times10^{-2}$ and $1.70\times10^{-2}$.
We can observe that among these unconstrained data-driven PF solvers, nonlinear PF solvers (i.e. the SVR and MLPNN methods) outperform the LR-based one, and the MLPNN method has better scalability than the SVR method.
Compare with the counterpart MLPNN method, our three physics-guided neural networks have more superior performance.
The average error of voltage solutions produced by our physics-guided neural networks is $4.57\times10^{-3}$ and about 4 times smaller than the MLPNN method.
This indicates that by including the reconstruction errors of ``decoders'' as an auxiliary task (i.e. PF modeling),
the proposed methods are capable of regularizing the MLPNN ``encoder'' and achieve better performance.
Besides, the cross-comparison of the three proposed methods shows that the more prior knowledge (i.e. physical laws) encoded, the more improvement the primary PF solving task achieves.
Fig. \ref{fig:inverse_CDF} shows the distribution of MAPE of bus voltages produced by the data-driven PF solvers on all the testing samples. 
90\% of testing errors of the LR, SVR, MLP, and our physics-guided NN methods are less than 32.39\%, 20.03\%, 19.16\%, and 5.23\%. 
It is clear that the unconstrained data-driven PF solvers have very bad solutions at some testing points, and our physics-guided data-driven PF solvers have a more stable performance on all the testing points. 

\subsubsection{Interpolation and Extrapolation Capability }
Power systems are dynamic, the load demands change significantly over time, and historical measurements may not observe the whole system dynamics.
Therefore, it is important to evaluate the generalization performance of data-driven PF solvers when given new power-flow inputs deviating a lot from training measurement data.
In this testing case, we normalize the measurement data values to a range of $[-1, 1]$ and divide these values equally into 20 portions.
We split the whole measurement data such that the values of input variables in training samples and testing samples are within different ranges. 
Then, we evaluate the interpolation and extrapolation capabilities of the aforementioned data-driven PF solvers. 

The RMSEs of interpolation and extrapolation results in each portion are illustrated in Fig. \ref{fig:inter_extra_polation}. 
The points in the shade areas are validation errors, and the points outside the shade areas are testing errors. 
As for the interpolation, 
we can see that it is difficult for the LR method to find a suitable hyperplane to fit the training data as well as to generate well on the testing data.
Although other methods fit the training data better than LR, the average differences between the testing errors and validation errors of the SVR, MLPNN and our physics-guided NN methods are 0.0185, 0.0145, and 0.00712 respectively. 
The interpolation capabilities of our physics-guided PF solvers are about 2 times better than the SVR and MLPNN methods.
As for the extrapolation, 
the LR method over-fits the training data and has the biggest testing errors. 
The average differences between the testing errors and validation errors of the SVR, MLPNN and our three physics-guided NN methods are 0.0125, 0.0122, and 0.00685 respectively. 
The extrapolation capabilities of our physics-guided PF solvers are about 2 times better than the SVR and MLPNN methods.
All these results show that with the constraints of the physical regularization, the generalization capabilities of our physics-guided PF solvers are at least 2 times better than those of other unconstrained PF solvers.

\subsubsection{Robustness to Data Outliers}
\begin{figure}[t]
    \centering
    \vspace{-4mm}
    \includegraphics[width=0.4\textwidth]{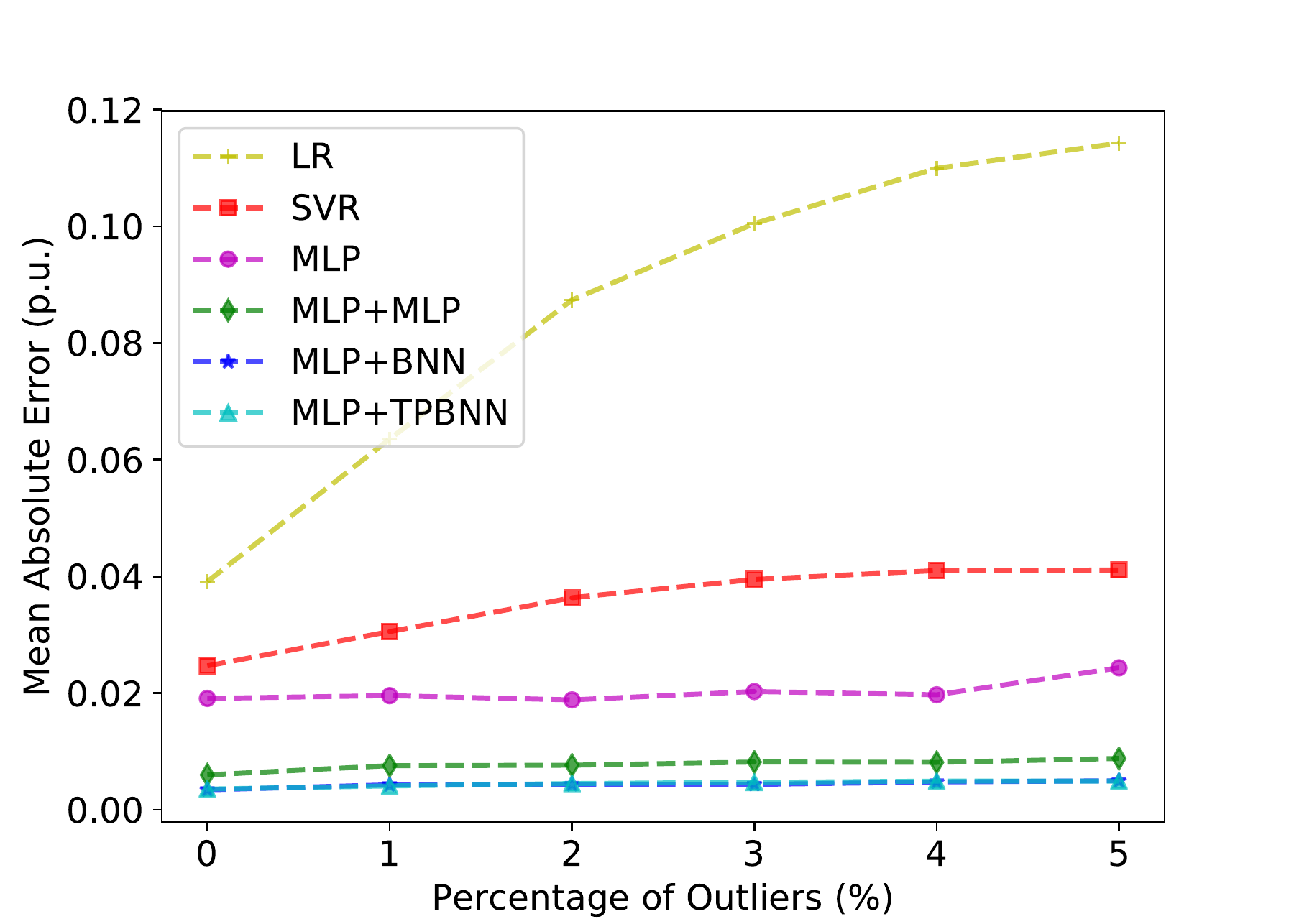}
    \vspace{-2mm}
    \caption{Voltage calculation accuracy of data-driven power flow solvers under different levels of outliers in training data.}
    \vspace{-4mm}
    \label{fig:118_inverse_outlier}
\end{figure}
It is common that measurement data is contaminated by noisy signals.
Therefore, we test the robustness of the data-driven PF solvers to noise and outliers in training data on the IEEE 118 bus system. 
In this test, different amounts of training data are corrupted with Gaussian noise, the standard deviation of which is $10\times IQR$ (Interquartile Range) of the training samples. 
The percentage of outliers resulting from this data corruption ranges from 0 to 10\%. 

The performance of estimation accuracy under different levels of outliers is illustrated in Fig. \ref{fig:118_inverse_outlier}.
The MAE (Mean Absolute Error) of the linear regression method increases fast when the percentage of outliers in the training data increases. 
This shows that linear regression is sensitive to the presence of outliers. 
On the other hand, the performances of the SVR and neural network methods are relatively stable when the level of outliers increases. 
Especially, our physics-guided neural networks continuously perform best and are robust enough even for 10\% outliers in training data. 
\begin{figure*}[!t]
\centering
\vspace{-6mm}
\subfloat[Admittance matrix $\boldsymbol{G}$]{\includegraphics[width=0.26\textwidth]{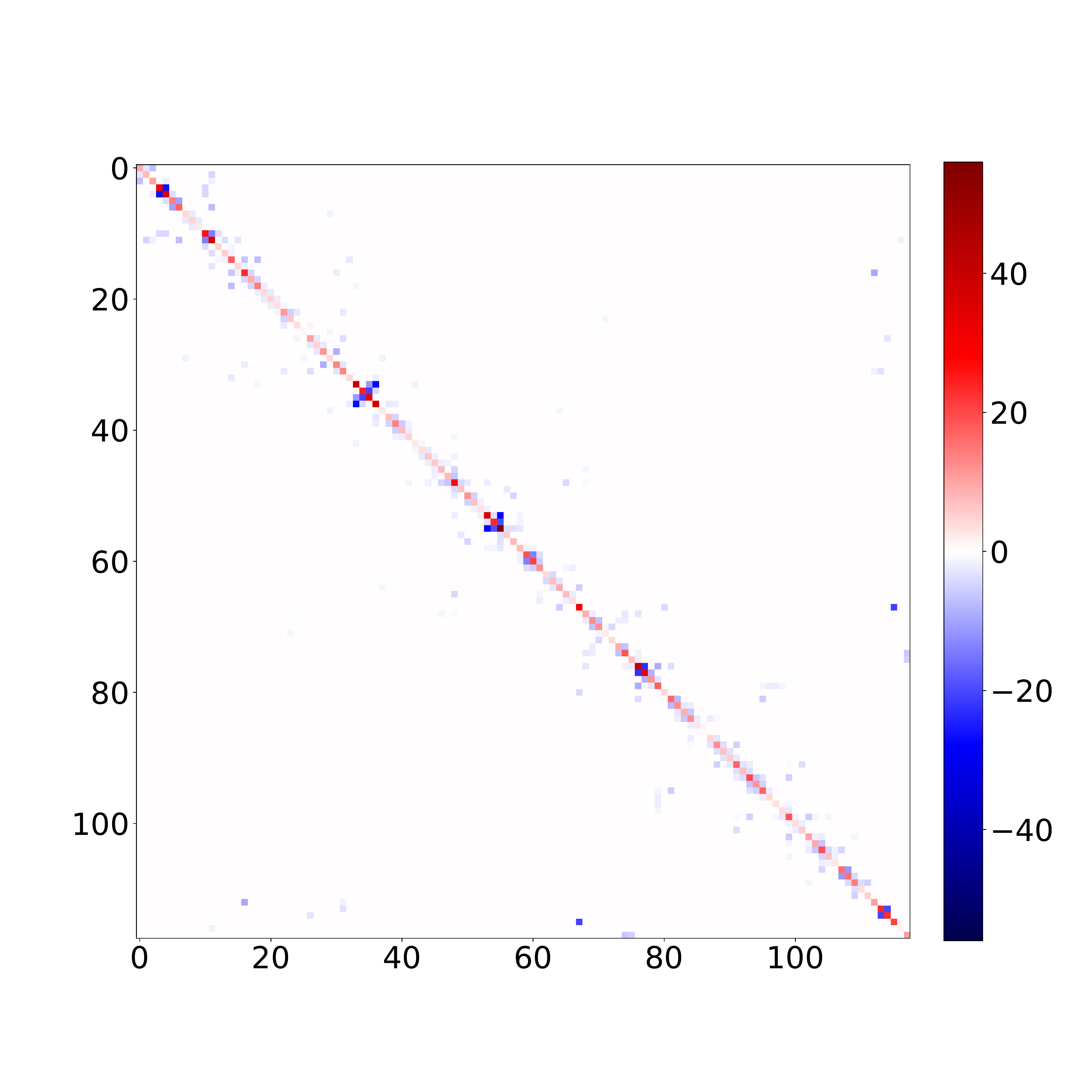}%
}
\hfill
\subfloat[BNN parameter matrix $\boldsymbol{W_G}$]{\includegraphics[width=0.26\textwidth]{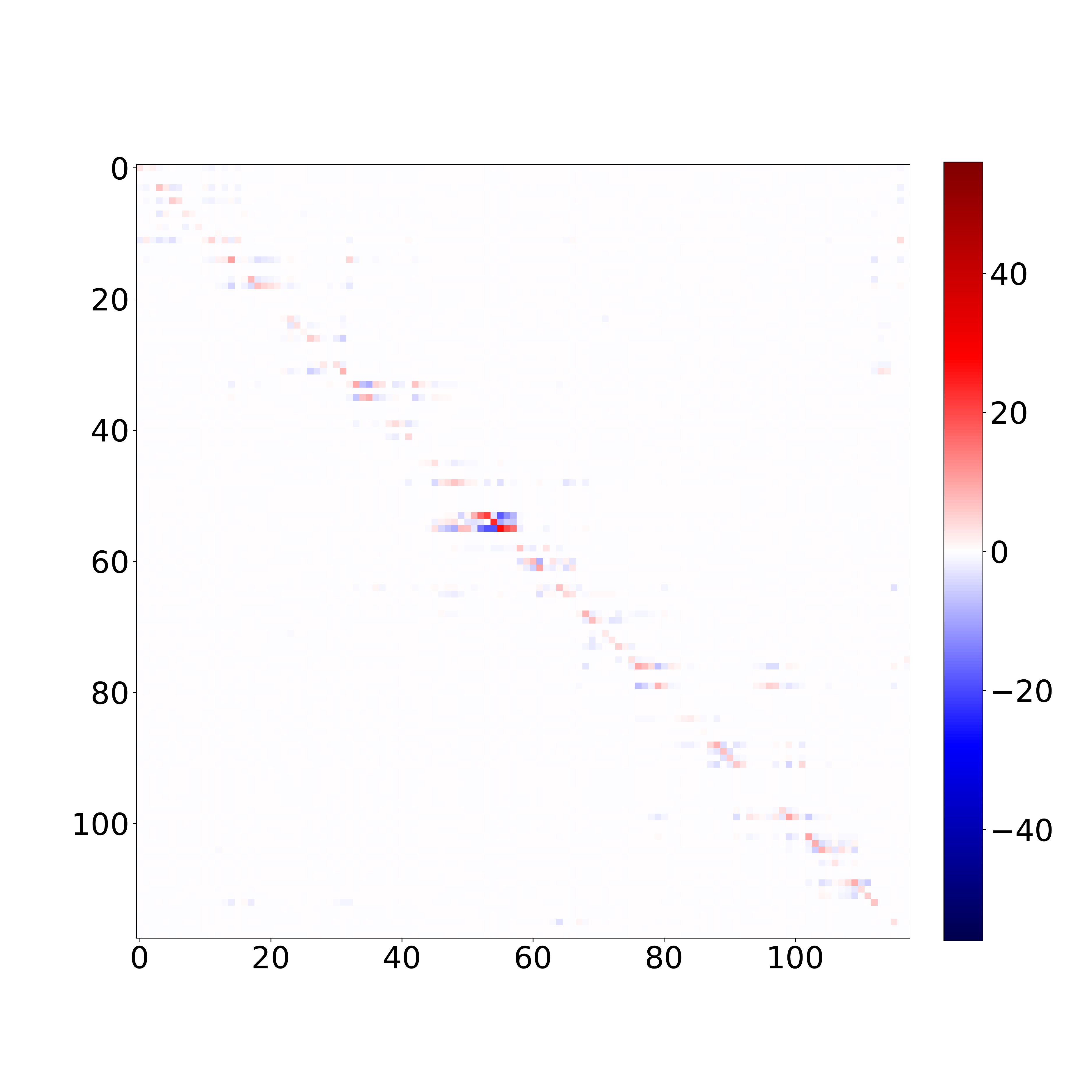}%
}
\hfill
\subfloat[TPBNN parameter matrix $\boldsymbol{W_G}$]{\includegraphics[width=0.26\textwidth]{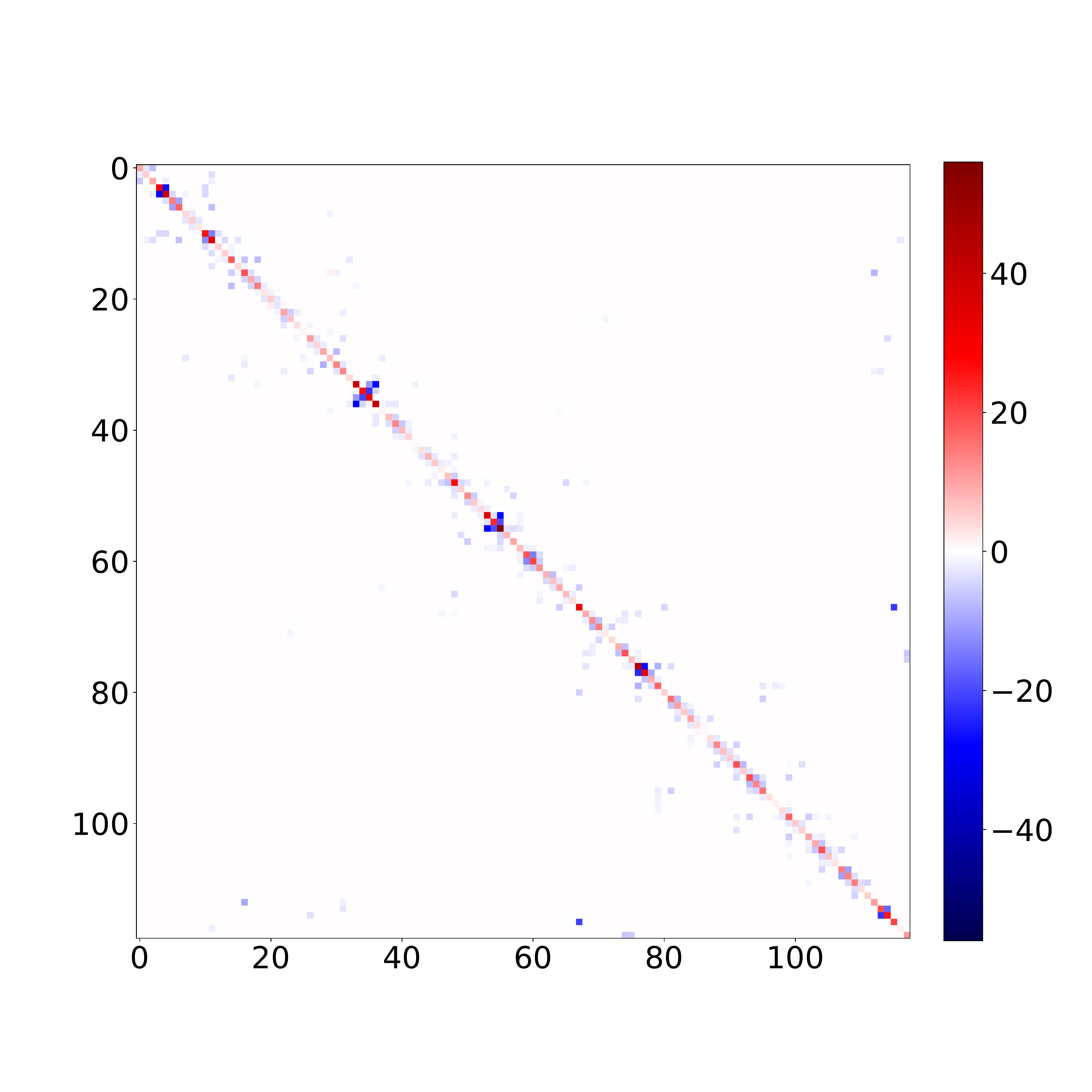}%
}
\caption{Comparisons between neural network parameter matrices and power system bus admittance matrices of IEEE 118-bus system.}
\vspace{-2mm}
\label{fig:118_GB_comparison}
\end{figure*}

\begin{table*}
    \vspace{-2mm}
    \caption{Power Flow Modeling Average Results}
    \centering
    \begin{tabular}{c|c||c|c|c|c|c}
\hline
\multicolumn{2}{c||}{ } & \multicolumn{5}{|c}{Root Mean Squared Error (p.u.)} \\
 \hline
 Test Case& Power &LR~\cite{liu2018data} & SVR~\cite{yu2017mapping} & MLP & BNN & TPBNN\\
 \hline
 \multirow{2}{*}{IEEE 57} & P & $3.36\times10^{-2}$ &$1.30\times10^{-2}$& $3.11\times10^{-2}\pm9.81\times10^{-4}$ & $8.50\times10^{-3}\pm1.74\times10^{-4}$ &$\mathbf{6.27\times10^{-4}\pm3.24\times10^{-6}}$\\
 & Q & $7.88\times10^{-2}$ &$6.08\times10^{-2}$& $2.51\times10^{-2}\pm1.95\times10^{-3}$ & $9.85\times10^{-3}\pm7.54\times10^{-5}$ &$\mathbf{1.53\times10^{-3}\pm7.76\times10^{-6}}$\\
 \hline
 \multirow{2}{*}{IEEE 118} & P & $4.53\times10^{-2}$ &$1.75\times10{-2}$& $3.82\times10^{-2}\pm3.05\times10^{-4}$ & $1.85\times10^{-2}\pm2.58\times10^{-4}$ &$\mathbf{3.13\times10^{-3}\pm1.13\times10^{-5}}$\\
 & Q & $1.96\times10^{-1}$ &$1.32\times10^{-1}$& $7.76\times10^{-2}\pm1.01\times10^{-2}$ & $6.76\times10^{-3}\pm1.25\times10^{-4}$ &$\mathbf{2.50\times10^{-3}\pm8.24\times10^{-6}}$\\
 \hline
\end{tabular}
    \vspace{-2mm}
    \label{tab:forward_meanae}
\end{table*}
\subsection{Physical Consistency Evaluation}
As shown in the architecture in Fig. \ref{fig:architecture}, our physics-guided NNs are also learning to approximate system parameters and rebuild PF models in their auxiliary tasks.
In this section, we evaluate the capabilities of our ``decoder'' parts to integrate physical knowledge and rebuild PF models.

First, the weight matrices of our BNN and TPBNN ``decoders'' are related to the bus admittance matrices $\boldsymbol{G}$ and $\boldsymbol{B}$. 
Fig. \ref{fig:118_GB_comparison} visualizes the bus admittance matrix $\boldsymbol{G}$ and the corresponding weight matrices of BNN and TPBNN under the IEEE 118 bus system. 
We observe similar results of the admittance matrix $\boldsymbol{B}$ approximation, but we do not show them due to page limitation.
Both BNN and TPBNN learn the sparsity of bus admittance matrices. 
Although BNN learns the diagonal non-zero parameters not as well as TPBNN, it roughly captures the power system topology through measurement data.
This means network topology is latently encoded in  system operation measurement data ~\cite{yu2017patopa, yu2017data}, and the 2nd-polynomial structure can help infer the topology pattern through data.
With the prior knowledge of topology, TPBNN provides a very good approximation of bus admittance matrices.

Regarding the performance of PF modeling, we compare our BNN, TPBNN, and MLPNN ``decoders'' with the LR~\cite{liu2018data} and SVR~\cite{yu2017mapping} methods which are also used to learn the mapping rules from bus voltages to power injections.
Given voltage measurements as inputs, the power injection modeling results of all these data-driven PF models are measured in terms of RMSE, shown in Table \ref{tab:forward_meanae}. 
The average errors of the LR, SVR, and our MLP, BNN, and TPBNN ``decoder'' methods on the two test systems are $8.84\times10^{-2}$, $5.58\times10^{-2}$, $4.30\times10^{-2}$, $1.09\times10^{-2}$, and $1.95\times10^{-3}$. 
The LR method has the worst performance, which indicates that linear regression has limited capabilities to represent nonlinear and complicated systems. 
The SVR method and our MLPNN ``decoder'' have comparable accuracy performance, but 
MLPNN scales better than SVR when system complexity increases.
Explicitly utilizing the good scalability of neural networks and the bilinear structural knowledge indicated by Kirchhoff’s laws, our well-designed BNN is better at capturing nonlinear dynamics of power systems.
Since ACPF equations are sparse polynomials, when the system topology is available, topology augmented BNN (i.e. TPBNN) prunes parameters, pays attention to a subset of its features and parameters, and consistently achieves the best accuracy performance than other methods.

\section{Discussion}

The proposed physics-guided neural networks enable the PF analysis in situations where the exact system parameters and control logic are difficult to obtain.
While our three kinds of physics-guided data-driven PF solvers achieve comparable calculation accuracy, they work in different situations according to the different prior physical knowledge of power systems.
The ``MLP+MLP'' model is a general method and especially suitable for power grids where some control policies are unknown due to the various active controllers introduced by the increasing penetration of DERs.
The ``MLP+BNN'' and ``MLP+TPBNN'' models are suitable for power grids where control policies indicate that a basic PF model is quadratic. 
Additionally, by utilizing the quadratic structural information, the BNN and TPBNN modules provide a new way to approximate system parameters and rebuild PF equations.  
However, TPBNN is a restricted method that requires accurate topology information.
In addition to PF analysis, since the calculation speed of a trained neural network is very fast, our physics-guided neural networks can apply to other power system problems requiring a large amount of PF calculations or real-time operations, such as probabilistic load flow analysis~\cite{fan2012probabilistic}, security constraints~\cite{gutierrez2010neural}, etc.

Although our physics-guided neural network methods solve the PF problem when PF models are inaccurate or unavailable, our methods cannot settle this matter once and for all.
Power grids are varying and evolving over time.
The well-trained neural network PF solvers need to adapt to future system changes.
One intuitive method is to periodically train the neural networks with new measurement data in an off-line manner and update the power-flow solvers with the newest neural network models. 
Some more advanced methods are to learn neural networks on the fly~\cite{sahoo2017online}, and to utilize time-series patterns of power systems~\cite{zhang2019real}, etc. 
Another direction of future work is to incorporate other generic physical knowledge and constraints that are helpful for the PF analysis, such as the symmetric structure of admittance matrix, inequality constraints (e.g., reactive power limits), etc.

\section{Conclusion}
To account for the increasing dynamics and uncertainties of modern power systems,
which make conventional PF analysis approaches ineffective,
researchers have proposed various data-driven methods to rebuild and solve power flow models through historical measurements.
However, existing works make invalid assumptions such as availability of power flow models and suffer from poor performance and generalizability.
We, therefore, propose physics-guided neural networks for PF analysis which incorporate physical constraints with practical assumptions.
Simulation results show that the proposed data-driven power flow solvers are regularized by corresponding physical laws and achieve higher accuracy and better physical consistency.
Besides, the results show that the algorithms (BNN and TPBNN), which incorporate the physical laws of the quadratic structure of ACPF equations and system topology, are promising alternatives to approximate system parameters and rebuild power flow equations.

\ifCLASSOPTIONcaptionsoff
  \newpage
\fi



%
\bibliographystyle{IEEEtran}
\bibliography{reference}

\begin{thebibliography}{10}
\providecommand{\url}[1]{#1}
\csname url@samestyle\endcsname
\providecommand{\newblock}{\relax}
\providecommand{\bibinfo}[2]{#2}
\providecommand{\BIBentrySTDinterwordspacing}{\spaceskip=0pt\relax}
\providecommand{\BIBentryALTinterwordstretchfactor}{4}
\providecommand{\BIBentryALTinterwordspacing}{\spaceskip=\fontdimen2\font plus
\BIBentryALTinterwordstretchfactor\fontdimen3\font minus
  \fontdimen4\font\relax}
\providecommand{\BIBforeignlanguage}[2]{{%
\expandafter\ifx\csname l@#1\endcsname\relax
\typeout{** WARNING: IEEEtran.bst: No hyphenation pattern has been}%
\typeout{** loaded for the language `#1'. Using the pattern for}%
\typeout{** the default language instead.}%
\else
\language=\csname l@#1\endcsname
\fi
#2}}
\providecommand{\BIBdecl}{\relax}
\BIBdecl

\bibitem{saadat1999power}
H.~Saadat, \emph{Power system analysis}.\hskip 1em plus 0.5em minus 0.4em\relax
  WCB/McGraw-Hill Singapore, 1999.

\bibitem{arrillaga1998ac}
J.~Arrillaga, J.~Arrillaga, and B.~Smith, \emph{AC-DC power system
  analysis}.\hskip 1em plus 0.5em minus 0.4em\relax IET, 1998, no.~27.

\bibitem{momoh1999review}
J.~A. Momoh, M.~El-Hawary, and R.~Adapa, ``A review of selected optimal power
  flow literature to 1993. ii. newton, linear programming and interior point
  methods,'' \emph{IEEE Transactions on Power Systems}, vol.~14, no.~1, pp.
  105--111, 1999.

\bibitem{chen2015measurement}
Y.~Chen, ``Measurement-based tools for power system monitoring and
  operations,'' 2015.

\bibitem{yu2017mapping}
J.~Yu, Y.~Weng, and R.~Rajagopal, ``Mapping rule estimation for power flow
  analysis in distribution grids,'' \emph{arXiv:1702.07948}, 2017.

\bibitem{sexauer2013phasor}
J.~Sexauer, P.~Javanbakht, and S.~Mohagheghi, ``Phasor measurement units for
  the distribution grid: Necessity and benefits,'' in \emph{Innovative Smart
  Grid Technologies (ISGT), 2013 IEEE PES}.\hskip 1em plus 0.5em minus
  0.4em\relax IEEE, 2013, pp. 1--6.

\bibitem{guttromson2002modeling}
R.~T. Guttromson, ``Modeling distributed energy resource dynamics on the
  transmission system,'' \emph{IEEE Transactions on Power Systems}, vol.~17,
  no.~4, pp. 1148--1153, 2002.

\bibitem{liscouski2004final}
B.~Liscouski and W.~Elliot, ``Final report on the august 14, 2003 blackout in
  the united states and canada: Causes and recommendations,'' \emph{A report to
  US Department of Energy}, vol.~40, no.~4, p.~86, 2004.

\bibitem{ferc2012arizona}
N.~Ferc, ``Arizona-southern california outages on 8 september 2011: causes and
  recommendations,'' \emph{FERC and NERC}, 2012.

\bibitem{yu2017data}
J.~Yu, Y.~Weng, and R.~Rajagopal, ``Data-driven joint topology and line
  parameter estimation for renewable integration,'' in \emph{2017 IEEE Power \&
  Energy Society General Meeting}.\hskip 1em plus 0.5em minus 0.4em\relax IEEE,
  2017, pp. 1--5.

\bibitem{yuan2016inverse}
Y.~Yuan, O.~Ardakanian, S.~Low, and C.~Tomlin, ``On the inverse power flow
  problem,'' \emph{arXiv preprint arXiv:1610.06631}, 2016.

\bibitem{yu2017patopa}
J.~Yu, Y.~Weng, and R.~Rajagopal, ``Patopa: A data-driven parameter and
  topology joint estimation framework in distribution grids,'' \emph{IEEE
  Transactions on Power Systems}, vol.~33, no.~4, pp. 4335--4347, 2017.

\bibitem{chen2016measurement}
Y.~C. Chen, J.~Wang, A.~D. Dom{\'\i}nguez-Garc{\'\i}a, and P.~W. Sauer,
  ``Measurement-based estimation of the power flow jacobian matrix,''
  \emph{IEEE Transactions on Smart Grid}, vol.~7, no.~5, pp. 2507--2515, 2016.

\bibitem{chen2013measurement}
Y.~C. Chen, A.~D. Dom{\'\i}nguez-Garc{\'\i}a, and P.~W. Sauer,
  ``Measurement-based estimation of linear sensitivity distribution factors and
  applications,'' \emph{IEEE Transactions on Power Systems}, vol.~29, no.~3,
  pp. 1372--1382, 2013.

\bibitem{liu2018data}
Y.~Liu, N.~Zhang, Y.~Wang, J.~Yang, and C.~Kang, ``Data-driven power flow
  linearization: A regression approach,'' \emph{IEEE Transactions on Smart
  Grid}, 2018.

\bibitem{karami2008radial}
A.~Karami and M.~Mohammadi, ``Radial basis function neural network for power
  system load-flow,'' \emph{International Journal of Electrical Power \& Energy
  Systems}, vol.~30, no.~1, pp. 60--66, 2008.

\bibitem{muller2010artificial}
H.~H. M{\"u}ller, M.~J. Rider, and C.~A. Castro, ``Artificial neural networks
  for load flow and external equivalents studies,'' \emph{Electric Power
  Systems Research}, vol.~80, no.~9, pp. 1033--1041, 2010.

\bibitem{baghaee2017three}
H.~R. Baghaee, M.~Mirsalim, G.~B. Gharehpetian, and H.~A. Talebi, ``Three-phase
  ac/dc power-flow for balanced/unbalanced microgrids including wind/solar,
  droop-controlled and electronically-coupled distributed energy resources
  using radial basis function neural networks,'' \emph{IET Power Electronics},
  vol.~10, no.~3, pp. 313--328, 2017.

\bibitem{nikkhajoei2006steady}
H.~Nikkhajoei and R.~Iravani, ``Steady-state model and power flow analysis of
  electronically-coupled distributed resource units,'' \emph{IEEE Transactions
  on Power Delivery}, vol.~22, no.~1, pp. 721--728, 2006.

\bibitem{gogna2016semi}
A.~Gogna and A.~Majumdar, ``Semi supervised autoencoder,'' in
  \emph{International Conference on Neural Information Processing}.\hskip 1em
  plus 0.5em minus 0.4em\relax Springer, 2016, pp. 82--89.

\bibitem{le2018supervised}
L.~Le, A.~Patterson, and M.~White, ``Supervised autoencoders: Improving
  generalization performance with unsupervised regularizers,'' in
  \emph{Advances in Neural Information Processing Systems}, 2018, pp. 107--117.

\bibitem{zamzam2019physics}
A.~S. Zamzam and N.~D. Sidiropoulos, ``Physics-aware neural networks for
  distribution system state estimation,'' \emph{arXiv preprint
  arXiv:1903.09669}, 2019.

\bibitem{andersson2008modelling}
G.~Andersson, ``Modelling and analysis of electric power systems,'' \emph{ETH
  Zurich}, pp. 5--6, 2008.

\bibitem{arulampalam2004control}
A.~Arulampalam, M.~Barnes*, A.~Engler, A.~Goodwin, and N.~Jenkins, ``Control of
  power electronic interfaces in distributed generation microgrids,''
  \emph{International Journal of Electronics}, vol.~91, no.~9, pp. 503--523,
  2004.

\bibitem{baker2019workshop}
N.~Baker, F.~Alexander, T.~Bremer, A.~Hagberg, Y.~Kevrekidis, H.~Najm,
  M.~Parashar, A.~Patra, J.~Sethian, S.~Wild \emph{et~al.}, ``Workshop report
  on basic research needs for scientific machine learning: Core technologies
  for artificial intelligence,'' USDOE Office of Science (SC), Washington, DC
  (United States), Tech. Rep., 2019.

\bibitem{maurer2006bounds}
A.~Maurer, ``Bounds for linear multi-task learning,'' \emph{Journal of Machine
  Learning Research}, vol.~7, no. Jan, pp. 117--139, 2006.

\bibitem{maurer2016benefit}
A.~Maurer, M.~Pontil, and B.~Romera-Paredes, ``The benefit of multitask
  representation learning,'' \emph{The Journal of Machine Learning Research},
  vol.~17, no.~1, pp. 2853--2884, 2016.

\bibitem{liu2016algorithm}
T.~Liu, D.~Tao, M.~Song, and S.~J. Maybank, ``Algorithm-dependent
  generalization bounds for multi-task learning,'' \emph{IEEE transactions on
  pattern analysis and machine intelligence}, vol.~39, no.~2, pp. 227--241,
  2016.

\bibitem{hong2014global}
T.~Hong, P.~Pinson, and S.~Fan, ``Global energy forecasting competition 2012,''
  2014.

\bibitem{zimmerman2010matpower}
R.~D. Zimmerman, C.~E. Murillo-S{\'a}nchez, and R.~J. Thomas, ``Matpower:
  Steady-state operations, planning, and analysis tools for power systems
  research and education,'' \emph{IEEE Transactions on power systems}, vol.~26,
  no.~1, pp. 12--19, 2010.

\bibitem{fan2012probabilistic}
M.~Fan, V.~Vittal, G.~T. Heydt, and R.~Ayyanar, ``Probabilistic power flow
  studies for transmission systems with photovoltaic generation using
  cumulants,'' \emph{IEEE Transactions on Power Systems}, 2012.

\bibitem{gutierrez2010neural}
V.~J. Gutierrez-Martinez, C.~A. Ca{\~n}izares, C.~R. Fuerte-Esquivel,
  A.~Pizano-Martinez, and X.~Gu, ``Neural-network security-boundary constrained
  optimal power flow,'' \emph{IEEE Transactions on Power Systems}, vol.~26,
  no.~1, pp. 63--72, 2010.

\bibitem{sahoo2017online}
D.~Sahoo, Q.~Pham, J.~Lu, and S.~C. Hoi, ``Online deep learning: Learning deep
  neural networks on the fly,'' \emph{arXiv preprint arXiv:1711.03705}, 2017.

\bibitem{zhang2019real}
L.~Zhang, G.~Wang, and G.~B. Giannakis, ``Real-time power system state
  estimation and forecasting via deep unrolled neural networks,'' \emph{IEEE
  Transactions on Signal Processing}, vol.~67, no.~15, pp. 4069--4077, 2019.

\end{thebibliography}

%








\end{document}